\documentclass[sigconf, screen]{acmart}
\usepackage{xcolor} 
\usepackage{multirow}
\usepackage{booktabs} 
\usepackage{amsmath} 
\usepackage{graphicx}
\usepackage[belowskip=-12pt]{caption}
\usepackage{rotating}
\usepackage{tabularx}
\usepackage{subfig}
\usepackage{wrapfig}
\usepackage{listings}
\usepackage{color, colortbl}
\usepackage{balance} 
\usepackage{lscape}
\usepackage{hyperref}
\usepackage{xspace}
\usepackage{framed}
\usepackage{syntax}
\usepackage{soul}
\usepackage{flushend}
   
\usepackage[tikz]{bclogo}
\usepackage{natbib}

\usepackage{hhline}
\usepackage{diagbox}
\usepackage{array, makecell}
\usepackage{microtype}

\usepackage{enumitem}
\definecolor{cellgreen}{RGB}{202,236,193}
\definecolor{cellblue}{RGB}{204, 229, 255}
\definecolor{cellpurple}{RGB}{216,191,216}
\definecolor{cellred}{RGB}{255, 204, 204}
\definecolor{cellorange}{RGB}{255, 229, 204}
\definecolor{cellgold}{RGB}{240,230,140}
\definecolor{cellsteel}{RGB}{230,230,250}

\definecolor{codegreen}{rgb}{0,0.6,0}
\definecolor{codegray}{rgb}{0.5,0.5,0.5}
\definecolor{codepurple}{rgb}{0.58,0,0.82}
\definecolor{backcolour}{rgb}{0.95,0.95,0.92}
\definecolor{Gray}{gray}{0.1}

\lstdefinestyle{mystyle}{
	backgroundcolor=\color{yellow!3}, 
	commentstyle=\color{codegreen}, 
	keywordstyle=\color{magenta},
	numberstyle=\tiny\color{codegray},
	stringstyle=\color{codegreen},
	basicstyle=\scriptsize,
	breakatwhitespace=false,         
	breaklines=true,                 
	captionpos=b,                    
	keepspaces=true,                 
	numbers=left,                    
	numbersep=5pt,                  
	showspaces=false,                
	showstringspaces=false,
	showtabs=false,                  
	tabsize=2,
	otherkeywords={self},
	frame=trbl,
	rulecolor=\color{black!15},
	commentstyle=\color{commentcolour}\slshape,
	upquote=true,
	xleftmargin=.015\textwidth, 
	xrightmargin=.005\textwidth
}

\lstdefinelanguage{Pythonna}{%
	language     = python,
	morekeywords = {to_categorical, flow_from_directory, pad_sequences, load_image}
}

\lstdefinestyle{customc}{
	belowcaptionskip=1\baselineskip,
	breaklines=false,
	frame= single,
	breaklines = true,
	xleftmargin=\parindent,
	language= Pythonna,
	showstringspaces=false,
	basicstyle=\footnotesize\ttfamily,
	keywordstyle=\bfseries\color{green!40!black},
	commentstyle=\itshape\color{purple!40!black},
	identifierstyle=\color{blue},
	stringstyle=\color{codegreen},
	backgroundcolor=\color{gray!4}
}

\graphicspath{{./figures/}}

\newcommand{\tabref}[1]{Table~\ref{#1}}
\newcommand{\fignref}[1]{Figure~\ref{#1}}

\newcommand{\secref}[1]{\S\ref{#1}}
\newcommand{\secnref}[1]{\S\ref{#1}}

\newcommand{\etal}{{\em et al.}\xspace}

\newcommand{\gc}{\textit{German Credit}\xspace}
\newcommand{\ac}{\textit{Adult Census}\xspace}
\newcommand{\bm}{\textit{Bank Marketing}\xspace}
\newcommand{\ti}{\textit{Titanic}\xspace}
\newcommand{\hc}{\textit{Home Credit}\xspace}
\newcommand{\cp}{\textit{Compas}\xspace}

\newcounter{rqs}
\stepcounter{rqs}

\newcounter{NumObservations}
\stepcounter{NumObservations}
\definecolor{shadecolor}{rgb}{.9,.9,.9}

\newcommand{\finding}[1]{%
	\begin{mdframed}[
		backgroundcolor= yellow!10,
		linewidth=.2pt,
		linecolor = gray!110,
		roundcorner=2pt,
		skipabove=6pt,
		skipbelow=-1pt
		innertopmargin=7pt,
		innerbottommargin=2pt,
		innerrightmargin=4pt,
		innerleftmargin=2pt,
		leftmargin = 0pt,
		rightmargin = 0pt]%
		\noindent{\textbf{Finding \arabic{NumObservations}}: \em #1} 
	\end{mdframed}%
	\stepcounter{NumObservations}
}

\AtBeginDocument{%
  \providecommand\BibTeX{{%
    \normalfont B\kern-0.5em{\scshape i\kern-0.25em b}\kern-0.8em\TeX}}}

\setcopyright{rightsretained}
\acmPrice{}
\acmDOI{10.1145/3468264.3468536}
\acmYear{2021}
\copyrightyear{2021}
\acmSubmissionID{fse21main-p52-p}
\acmISBN{978-1-4503-8562-6/21/08}
\acmConference[ESEC/FSE '21]{Proceedings of the 29th ACM Joint European Software Engineering Conference and Symposium on the Foundations of Software Engineering}{August 23--28, 2021}{Athens, Greece}
\acmBooktitle{Proceedings of the 29th ACM Joint European Software Engineering Conference and Symposium on the Foundations of Software Engineering (ESEC/FSE '21), August 23--28, 2021, Athens, Greece}

\begin{document}

\title{Fair Preprocessing: Towards Understanding Compositional Fairness of Data Transformers in Machine Learning Pipeline}

\author{Sumon Biswas}
\affiliation{
	\institution{Dept. of Computer Science, Iowa State University}
	\city{Ames}
	\state{IA}
	\country{USA}}
\email{sumon@iastate.edu}

\author{Hridesh Rajan}
\affiliation{
	\institution{Dept. of Computer Science, Iowa State University}
	\city{Ames}
	\state{IA}
	\country{USA}}
\email{hridesh@iastate.edu}

\begin{abstract}

In recent years, many incidents have been reported where machine learning models exhibited discrimination among people based on race, sex, age, etc. Research has been conducted to measure and mitigate unfairness in machine learning models. For a machine learning task, it is a common practice to build a pipeline that includes an ordered set of data preprocessing stages followed by a classifier. However, most of the research on fairness has considered a single classifier based prediction task. What are the fairness impacts of the preprocessing stages in machine learning pipeline? Furthermore, studies showed that often the root cause of unfairness is ingrained in the data itself, rather than the model. But no research has been conducted to measure the unfairness caused by a specific transformation made in the data preprocessing stage. In this paper, we introduced the causal method of fairness to reason about the fairness impact of data preprocessing stages in ML pipeline. We leveraged existing metrics to define the fairness measures of the stages. Then we conducted a detailed fairness evaluation of the preprocessing stages in 37 pipelines collected from three different sources. Our results show that certain data transformers are causing the model to exhibit unfairness. We identified a number of fairness patterns in several categories of data transformers. Finally, we showed how the local fairness of a preprocessing stage composes in the global fairness of the pipeline. We used the fairness composition to choose appropriate downstream transformer that mitigates unfairness in the machine learning pipeline.
\end{abstract}



\maketitle

\section{Introduction}
\label{sec:introduction}

Fairness of machine learning (ML) predictions is becoming more important with the rapid increase of ML software usage in important decision making \cite{dixon2018measuring, olson2011algorithm, angwin2016machine, goodall2016can}, and the black-box nature of ML algorithms \cite{galhotra2017fairness, aggarwal2019black}. There is a rich body of work on measuring fairness of ML models \cite{zafar2015fairness, dwork2012fairness, feldman2015certifying, hardt2016equality, calders2010three, chouldechova2017fair, zemel2013learning, speicher2018unified} and mitigate the bias \cite{zhang2018mitigating, calders2010three, zafar2015fairness, chouldechova2017fair, goh2016satisfying, hardt2016equality, pleiss2017fairness, kamiran2012decision}. Recent work \cite{brun2018software, holstein2019improving, friedler2019comparative, biswas20machine, harrison2020empirical, chakraborty2019software} has shown that more software engineering effort is required towards detecting bias in complex environment and support developers in building fairer models. 

The majority of work on ML fairness has focused on classification task with single classifier \cite{galhotra2017fairness, feldman2015certifying, aggarwal2019black, bower2017fair}. However, real-world machine learning software operate in a complex environment \cite{bower2017fair, d2020fairness}. In an ML task, the prediction is made after going through a series of stages such as data cleaning, feature engineering, etc., which build the machine learning pipeline \cite{amershi2019software, yang2020fairness}. Studying only the fairness of the classifiers (e.g., \textit{Decision Tree}, \textit{Logistic Regression}) fails to capture the fairness impact made by other stages in ML pipeline. In this paper, we conducted a detailed analysis on how the data preprocessing stages affect fairness in ML pipelines. 

Prior research observed that bias can be encoded in the data itself and missing the opportunity to detect bias in earlier stage of ML pipeline can make it difficult to achieve fairness algorithmically~\cite{kirkpatrick2017s, holstein2019improving, grgic2018beyond,dixon2018measuring}. Additionally, bias mitigation algorithms operating in the preprocessing stage were shown to be successful \cite{kamiran2012data, feldman2015certifying}. Therefore, it is evident that the preprocessing stages of ML pipeline can introduce bias. However, no study has been conducted to measure the fairness of the preprocessing stages and show how it impacts the overall fairness of the pipeline.
In this paper, we used the causal method of fairness to reason about the fairness impact of preprocessing stages in ML pipeline. Then, we leveraged existing fairness metrics to measure fairness of the preprocessing stages. Using the measures, we conducted a thorough analysis on a benchmark of 37 real-world ML pipelines collected from three different sources, which operate on five datasets.
These ML pipelines allowed us to evaluate fairness of a wide selection of preprocessing stages from different categories such as data standardization, feature selection, encoding, over/under-sampling, imputation, etc. For comparative analysis, we also collected data transformers e.g., \texttt{StandardScaler}, \texttt{MinMaxScaler}, \texttt{PCA}, \texttt{l1-normalizer}, \texttt{QuantileTransformer}, etc., from the pipelines as well as corresponding ML libraries, and evaluated fairness. Finally, we investigated how fairness of these preprocessing techniques (\textit{local fairness}) composes with other preprocessing stages, and the whole pipeline (\textit{global fairness}). Specifically, we answered the following three research questions.

\textbf{RQ1} (fairness of preprocessing stages): What are the fairness measures of each preprocessing stage in ML pipeline?
\textbf{RQ2} (fair transformers): What are the fair (and biased) data transformers among the commonly used ones? 
\textbf{RQ3} (fairness composition): How fairness of data preprocessing stages composes in ML pipeline?

\begin{itemize}
    \item How local fairness compose into global fairness?
    \item Does choosing a downstream transformer depend on the fairness of an upstream transformer?
\end{itemize}

To the best of our knowledge, we are the first to evaluate the fairness of preprocessing stages in ML pipeline. 
Our results show that by measuring the fairness impact of the stages, the developers would be able to build fairer predictions effectively.
Furthermore, the libraries can provide fairness monitoring into the data transformers, similar to the performance monitoring for the classifiers. Our evaluation on real-world ML pipelines also suggests opportunities to build automated tool to detect unfairness in the preprocessing stages, and instrument those stages to mitigate bias.
We have made the following contributions in this paper:

\begin{enumerate}[leftmargin=*]
    \item We created a fairness benchmark of ML pipelines with several stages. The benchmark, code and results are shared in our replication package\footnote{https://github.com/sumonbis/FairPreprocessing} in GitHub repository, that can be leveraged in further research on building fair ML pipeline.

    \item We introduced the notion of causality in ML pipeline and leveraged existing metrics to measure the fairness of preprocessing stages in ML pipeline.
    
    \item Unfairness patterns have been identified for a number of stages. 
    
    \item We identified alternative data transformers which can mitigate bias in the pipeline.
    
    \item Finally, we showed the composition of stage-specific fairness into overall fairness, which is used to choose appropriate downstream transformer that mitigates bias.
\end{enumerate}

The paper is organized as follows: \secref{sec:motivate} describes the motivating examples, \secref{sec:methodology} describes the existing metrics and our approach. In \secref{sec:evaluation}, we described the benchmark and experiments. \secref{sec:stages} explores the results, \secref{sec:transformers} provides a comparative study among transformers, and \secref{sec:composition} evaluates the fairness composition. Finally, \secref{sec:threats} describes the threats to validity, \secref{sec:related} discusses related work, and \secref{sec:conc} concludes.  
\section{Motivation}
\label{sec:motivate}

In this section, we present two ML pipelines which show that the preprocessing stage affects the fairness of the model and it is important to study the bias induced by certain data transformers.

\subsection{Motivating Example 1}
\label{subsec:motivate1}

Yang \etal \cite{yang2020fairness} studied the following ML pipeline which was originally outlined by Propublica for recidivism prediction on \cp dataset \cite{angwin2016machine}.
The goal is predict future crimes based on the data of defendants.
The fairness values, in terms of statistical parity difference (SPD: -0.102) and equal opportunity difference (EOD: -0.027), suggest that the prediction is biased towards\footnote{Bias \textit{towards} a group connotes that the prediction favours that group.} \textit{Caucasian} defendants when \textit{race} is considered as sensitive attribute. The pipeline consists of several preprocessing stages before applying \texttt{LogisticRegression} classifier. Data preprocessing includes cleaning, encoding categorical features, and missing value imputation. Recent research \cite{yang2020fairness} showed that the data transformation in this pipeline is not symmetric across gender groups i.e., male defendants are filtered more than the female.
Do these data transformations introduce unfairness in the prediction? If yes, what are the unfairness measures of these transformers? Is it possible to leverage existing metrics to measure the unfairness of each component? If we can understand the effect of each data transformer, it would be possible to choose data preprocessing technique wisely to avoid introducing bias as well as mitigate the inherent bias in data or classifier.

\begin{lstlisting}[language=Python]
df = pd.read_csv(f_path)
df = df[(df.days_b_screening_arrest <= 30)
    & (df.days_b_screening_arrest >= -30)
    & (df.is_recid != -1) & (df.c_charge_degree != 'O')
    & (df.score_text != 'N/A')]
df = df.replace('Medium', 'Low')
labels = LabelEncoder().fit_transform(df.score_text)
impute1_onehot = Pipeline([
    ('imputer1', SimpleImputer(strategy='most_frequent')),
    ('onehot', OneHotEncoder(handle_unknown='ignore'))])
impute2_bin = Pipeline([
    ('imputer2', SimpleImputer(strategy='mean')), 
    ('discretizer', KBinsDiscretizer(n_bins=4, encode='ordinal', strategy='uniform'))])
featurizer = ColumnTransformer(transformers=[
    ('impute1_onehot', impute1_onehot, ['is_recid']),
    ('impute2_bin', impute2_bin, ['age'])])
pipeline = Pipeline([('features', featurizer),
    ('classifier', LogisticRegression())])
\end{lstlisting}

\subsection{Motivating Example 2}
\label{subsec:motivate2}

The following ML pipeline is collected from the benchmark used by Biswas and Rajan \cite{biswas20machine} for studying fairness of ML models. This pipeline operates on \gc dataset. Here, the goal is to predict the credit risk (good/bad) of individuals based on their personal data such as age, sex, income, etc.
In this pipeline, before training the classifier, data has been processed using two transformers: \texttt{PCA} for principal component analysis, and \texttt{SelectKBest} for selecting high-scoring features. The fairness value (SPD: 0.005) shows that prediction is slightly biased towards \textit{female} candidates. However, if the transformers are not applied, then prediction becomes biased towards \textit{male} (SPD: -0.117).
By applying one transformer at a time, we observed that \texttt{PCA} alone is not causing the change of fairness. In this case, \texttt{SelectBest} is causing bias towards \textit{female}, which in turn mitigating the overall fairness of the pipeline. 
Therefore, in addition to study the fairness of transformers in isolation, it is important to understand how fairness of components composes in the pipeline.

\begin{lstlisting}[language=Python]
features = []
features.append(('pca', PCA(n_components=2)))
features.append(('select_best', SelectKBest(6)))
feature_union = FeatureUnion(features)
estimators = []
estimators.append(('feature_union', feature_union))
estimators.append(('RF', RandomForestClassifier()))
model = Pipeline(estimators)
model.fit(X_train, y_train)
y_pred = model.predict(X_test)
\end{lstlisting}

Our key idea is to leverage causal reasoning and observe fairness impact of a stage on prediction. To do that we create alternative pipeline by removing a stage.
For example, from the above pipeline, we remove the \texttt{SelectKBest} and compare the predictions with original pipeline. We observe that \texttt{SelectKBest} 
is causing 1.1\% of the female and 3.6\% of the male participants to change predictions from favorable (good credit) to unfavorable (bad credit). Since the stage is causing more unfavorable decisions to male, the stage is biased towards female. Thus, we used existing fairness criteria to measure fairness impact of a stage and propose novel metrics.
\section{Methodology}
\label{sec:methodology}

In this section, first, we describe the background of ML pipeline, focussing on the data preprocessing stages. Second, we formulate the method and metrics to measure fairness of a certain preprocessing stage with respect to the pipeline it is used within. 

\subsection{ML Pipeline}

\begin{figure}[]
	\centering
	\includegraphics[width=\columnwidth]{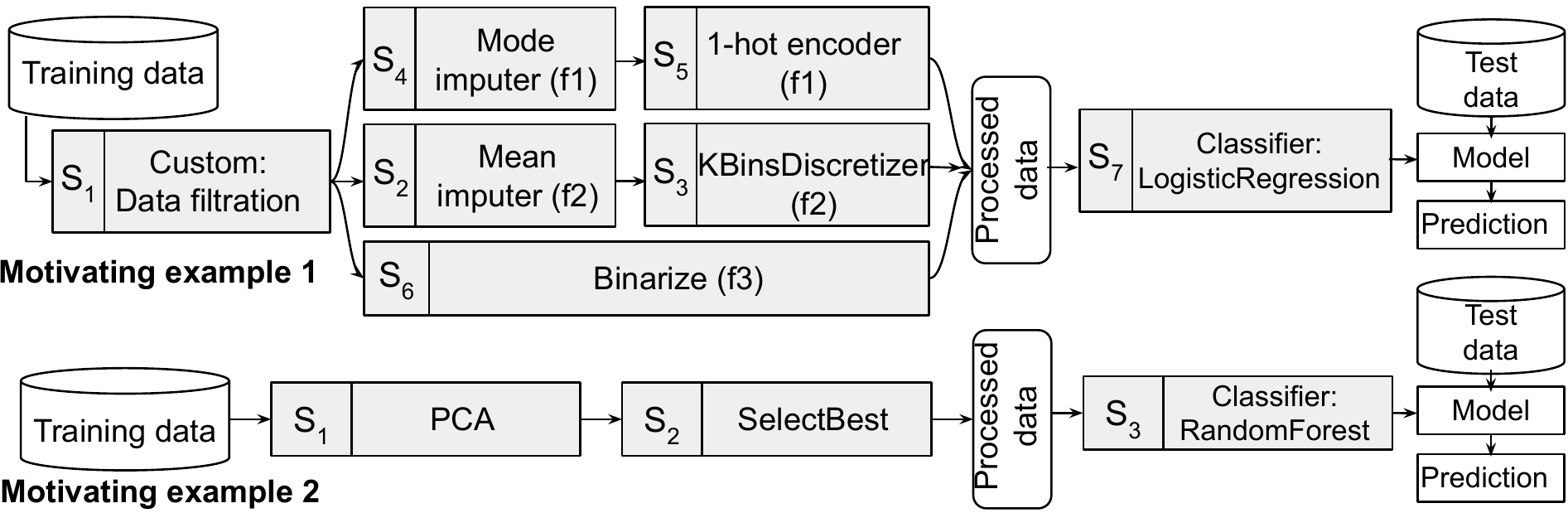}
	\caption{ML pipelines for the motivating examples, having a sequence of preprocessing stages followed by a classifier.}
	\label{canonical-pipeline}
\end{figure}

Amershi \etal proposed a nine-stage machine learning pipeline with data-oriented (collection, cleaning, and labeling) and model-oriented (model requirements, feature engineering, training, evaluation, deployment, and monitoring) stages \cite{amershi2019software}. 
Other research~\cite{abadi2016tensorflow, baylor2017tfx} also described data preprocessing as an integral part of the ML pipeline. 
The pipelines in the motivating examples are depicted in \fignref{canonical-pipeline}, which follows the representation provided by Yang \etal \cite{yang2020fairness}.
In this paper, we adapted the canonical definition of pipeline from Scikit-Learn pipeline specification \cite{buitinck2013api, sklearn-pipeline}, which is aligned with the ML models studied in the literature for fair classification tasks \cite{bellamy2018ai, yang2020fairness,biswas20machine, friedler2019comparative, aggarwal2019black, galhotra2017fairness}. 
We are interested in investigating the fairness of the data preprocessing stages in the pipeline, which is depicted with grey boxes in \fignref{canonical-pipeline}.

To summarize, a canonical \textit{ML pipeline} is an ordered set of $m$ stages with a set of \textit{preprocessing stages} ($S_1, S_2, \dots S_{m-1}$) and a final \textit{classifier} ($S_m$).
Each preprocessing stage, $S_k$ operates on the data already processed by preceding stages $S_1, \dots S_{k-1}$. A data preprocessing stage $S_k$ can be a data transformer or a set of custom operations. A \textit{data transformer} is a well-known algorithm or method to perform a specific operation such as variable encoding, feature selection, feature extraction, dimensionality reduction, etc. on the data \cite{buitinck2013api}. For example, in the second motivating example, two transformers (\texttt{PCA} and \texttt{SelectKBest}) have been used. \textit{Custom transformation} includes data/task-specific contextual operations on the dataset. For example, in \secref{subsec:motivate1} (line 2-3), the data instances that do not contain a value in the range [-30, 30] for the feature \texttt{days_b_screening_arrest}, have been filtered. This means the pipeline ignored the data of the defendants with more than 30 days between their screening and arrest. 
This formulation of ML pipeline allowed us to evaluate fairness of the preprocessing stages in real-world ML tasks.

\subsection{Existing Fairness Metrics}
\label{subsec:existing-metrics}

We have leveraged existing fairness metrics to measure the fairness of the whole pipeline. Many fairness metrics have been proposed in the literature for measuring fairness of classification tasks \cite{bellamy2018ai, binns2017fairness, dixon2018measuring}.
In general, the fairness metrics compute group-specific classification rates (e.g., true positives, false positives), and calculates the difference between groups to measure the fairness.
In this paper, we adopted the representative group fairness metrics used by \cite{friedler2019comparative, biswas20machine}.
Specifically, we leveraged the following metrics: statistical parity difference (SPD) \cite{kamiran2012data, zafar2015fairness, feldman2015certifying}, equal opportunity difference (EOD) \cite{hardt2016equality}, average odds difference (AOD) \cite{hardt2016equality}, and error rate difference (ERD) \cite{chouldechova2017fair}. 
Given a dataset $D$ with $n$ instances, let, actual classification label be $Y$, predicted classification label be $\hat{Y}$, and sensitive attribute be $\mathcal{A}$. Here, $Y = 1$ if the label is favorable to the individuals, otherwise $Y = 0$. For example, classification task on \gc dataset predicts the credit risk (good/bad credit) of individuals. In this case, $Y = 1$ if the prediction is \textit{good credit}, otherwise $Y = 0$. Suppose, for privileged group (e.g., \textit{White}), $\mathcal{A} = 1$ and for unprivileged group (e.g., \textit{non-White}), $\mathcal{A} = 0$.
SPD is computed by observing the probability of giving favorable label to each group and taking the difference. EOD measures the true-positive rate difference between groups. AOD calculates both true positive rate and false positive rate difference and then takes the average. ERD calculates the sum of false positive rate difference and false negative rate difference between groups. The definitions of these metrics are as follows:

{\footnotesize
\begin{align}
    \mathsf{SPD} &= \mathsf{P} [\hat{Y} = 1 | \mathcal{A} = 0] - \mathsf{P} [\hat{Y} = 1 | \mathcal{A} = 1] \nonumber
    \\
    \mathsf{EOD} &= \mathsf{P} [\hat{Y} = 1 | Y = 1, \mathcal{A} = 0] - \mathsf{P} [\hat{Y} = 1 | Y = 1, \mathcal{A} = 1] \nonumber
    \\
    \mathsf{AOD} &= (1/2) \{(\mathsf{P} [\hat{Y} = 1 | Y = 1, \mathcal{A} = 0] - \mathsf{P} [\hat{Y} = 1 | Y = 1, \mathcal{A} = 1]) \nonumber \\
    &+ (\mathsf{P} [\hat{Y} = 1 | Y = 0, \mathcal{A} = 0] - \mathsf{P} [\hat{Y} = 1 | Y = 0, \mathcal{A} = 1]) \} \nonumber
    \\
    \mathsf{ERD} &= (\mathsf{P} [\hat{Y} = 1 | Y = 0, \mathcal{A} = 0] - \mathsf{P} [\hat{Y} = 1 | Y = 0, \mathcal{A} = 1]) \} \nonumber \\
    &+ (\mathsf{P} [\hat{Y} = 0 | Y = 1, \mathcal{A} = 0] - \mathsf{P} [\hat{Y} = 0 | Y = 1, \mathcal{A} = 1]) \}
    \label{eq1}
\end{align}
}%

Disparate impact (DI) and statistical parity difference (SPD) both measure the same rate i.e., probability of classifying data instance as favorable, but DI computes the ratio of privileged and unprivileged groups' rate, whereas SPD computes the difference. Therefore, from DI and SPD, we only used SPD in our evaluation.

\subsection{Fairness of Preprocessing Stages}
\label{subsec:new-metrics}

Suppose, $\mathcal{P}$ is a pipeline with $m$ stages and our goal is to evaluate the fairness of the stage $S_k$, where $1 \le k < m$.
In other words, we want to measure the fairness impact of $S_k$ on the prediction made by $\mathcal{P}$. To achieve that we applied the causal reasoning for evaluating fairness. The causality theorem was proposed by \citeauthor{pearl2000causality}~\cite{pearl2000causality,pearl2009causal} and further studied extensively to reason about fairness in many scenarios \cite{galhotra2017fairness, kusner2017counterfactual, zhang2018fairness, salimi2019interventional, russell2017worlds}.
Causality notion of fairness captures that everything else being equal, the prediction would not be changed in the counterfactual world where only an intervention happens on a variable \cite{galhotra2017fairness, kusner2017counterfactual, russell2017worlds}. For example, \citeauthor{galhotra2017fairness} proposed causal discrimination score for fairness testing \cite{galhotra2017fairness}. The authors created test inputs by altering original protected attribute values of each data instance, and observed whether prediction is changed for those test inputs.
If the intervention causes the prediction to be changed, we call the software causally unfair with respect to that intervention. In our case, if a preprocessing stage $S_k$ be the intervention, to measure the fairness of $S_k$, we have to capture the prediction disparity caused by the intervention $S_k$. This causal reasoning of fairness is a stronger notion since it provides causality in software by observing changes in the outcome made by a specific stage in the pipeline \cite{galhotra2017fairness,pearl2009causal}.

\subsubsection{Causal Method to Measure Fairness of Preprocessing Stage}
From pipeline $\mathcal{P}$, we construct another pipeline $\mathcal{P^*}$ by only excluding the stage $S_k$ from $\mathcal{P}$. 
After applying the stage $S_k$ in $\mathcal{P^*}$, to what extent the prediction of $\mathcal{P^*}$ changes, and whether the change is favorable to any group?
Broadly, this can be measured by observing the prediction difference between $\mathcal{P}$ and $\mathcal{P^*}$ and computing the fairness of these changes using the fairness metrics from \eqref{eq1}.

Suppose, the predictions made by the two pipelines are $\hat{Y}(\mathcal{P})$ and $\hat{Y}(\mathcal{P^*})$. 
Let, $I$ be the impact set for $S_k$, which denotes the prediction parity between $\hat{Y}(\mathcal{P})$ and $\hat{Y}(\mathcal{P^*})$ such that 
for $i^{th}$ data instance, if $\hat{Y}_i(\mathcal{P}) = \hat{Y}_i(\mathcal{P^*})$, then $I_i = 0$, otherwise $1$. 
By causality, the fairness of preprocessing stage (denoted by $\mathsf{SF}$) is calculated based on $[\hat{Y}(\mathcal{P})$, $\hat{Y}(\mathcal{P^*})]_{I=1}$ with respect to a fairness metric $M$, which is shown in \eqref{eq2a}. We noticed that a few preprocessing stages, specifically the encoders can not be removed without replacing with an alternative stage. For such situations, we have defined the fairness of $S_k$ with reference to another stage $S_{k}'$, denoted by $\mathsf{SF (S_k|S_k')}$ in \eqref{eq2b}.

Zelaya also used the similar method for quantifying the effect of a preprocessing stage with a goal of computing \textit{volatility} of a stage \cite{zelaya2019towards}. \textit{Volatility} quantifies how much impact a preprocessing stage has on the outcome by computing the probability of prediction changes. However, it does not capture the fairness of the stage, since a stage can cause high change in the prediction by maintaining the predictions fair. Next in \secref{subsec:metrics}, we have extended our causality based formulation of \eqref{eq2a} for each fairness metric in \eqref{eq1} to capture the fairness impact of each preprocessing stage. Similar to \cite{galhotra2017fairness}, the benefit of this formulation is, the measures do not require an oracle, since the prediction equivalence of pipelines $\mathcal{P}$ and $\mathcal{P^*}$ serves the goal of evaluating fairness of the stage. Note that the rest of the definitions in \secref{subsec:metrics} are independent of \eqref{eq2a} and \eqref{eq2b}.

{\footnotesize
\begin{subequations}\label{eq2}
    \begin{gather}
        I = 
        \begin{cases}
            0 & \text{if } \hat{Y}_i(\mathcal{P}) = \hat{Y}_i(\mathcal{P^*}) \nonumber \\
            1 & \text{otherwise}        
        \end{cases} \text{, for all } i \in \{1 \dots n\} \nonumber \\
        \mathsf{SF}(S_k) = M[\hat{Y}(P), \hat{Y}(P^*)]_{I = 1} \text{ where } \mathcal{P^*} = \mathcal{P} \setminus S_k \label{eq2a}\\
        \mathsf{SF}(S_k|S_k') = M[\hat{Y}(P), \hat{Y}(P^*)]_{I = 1} \text{ where } \mathcal{P^*} = (\mathcal{P} \setminus S_k) \cup S_k' \label{eq2b}
    \end{gather}
\end{subequations}
}%

\subsubsection{Fairness Metircs for Preprocessing Stage}
\label{subsec:metrics}

We have leveraged the definition of metrics SPD, EOD, AOD, and ERD from \eqref{eq1} to capture the \underline{s}tage-specific \underline{f}airness of $S_k$. Essentially, the new metrics will identify the disparities between $\hat{Y}_i(\mathcal{P})$ and $\hat{Y}_i(\mathcal{P^*})$ and use corresponding fairness criteria to measure how much $S_k$ favors a specific group with respect to other group(s).

Suppose, among $n$ data instances, $n_u$ are from the unprivileged group and $n_p$ from the privileged group.
$\mathsf{SFC}_{\mathsf{SPD}}$ computes how many of the data instances have been changed from unfavorable to favorable after applying the stage $S_k$. 
To do that we count changes in both directions (unfavorable to favorable and favorable to unfavorable), and take the difference. The sign of $\mathsf{SFC}_{\mathsf{SPD}}$ preserves the direction of changes.
Finally, the metric $\mathsf{SF}_{\mathsf{SPD}}$ is computed by taking the difference of \underline{r}ates ($\mathsf{SFR}_{\mathsf{SPD}}$) between unprivileged and privileged groups. 
Note that the metric captures fairness by measuring the difference of favorable change rates between groups. Simply counting the mismatches between $\hat{Y}_i(\mathcal{P})$ and $\hat{Y}_i(\mathcal{P^*})$ could provide degree of changes in $\mathsf{SFC}_{\mathsf{SPD}}$ but would not capture fairness. Furthermore, computing favorable changes to both groups separately and evaluating the disparity between them captures fairness according to the original definition of $\mathsf{SPD}$.

{\footnotesize
\begin{gather}
    \mathsf{SFC}_{i\mathsf{SPD}} =
    \begin{cases}
        1 & \text{if } \hat{Y}_i(\mathcal{P}) = 1 \text{ and } \hat{Y}_i(\mathcal{P^*}) = 0 \\
        -1 & \text{if } \hat{Y}_i(\mathcal{P}) = 0 \text{ and } \hat{Y}_i(\mathcal{P^*}) = 1 \nonumber \\
        0 & \text{otherwise}
    \end{cases}
    \\
    \mathsf{SFC}_{\mathsf{SPD}} = \sum_{i=1}^{n} \mathsf{SFC}_{i{\mathsf{SPD}}} \nonumber 
    \\
    \mathsf{SFR}_{\mathsf{SPD}}(u) = \mathsf{SFC}_{\mathsf{SFD}}(u)/n_u \text{, } \mathsf{SFR}_{\mathsf{SPD}}(p) = \mathsf{SFC}_{\mathsf{SPD}}(p)/n_p \nonumber 
    \\
    \mathsf{SF}_{\mathsf{SPD}} = \mathsf{SFR}_{\mathsf{SPD}}(u) - \mathsf{SFR}_{\mathsf{SPD}}(p)
    \label{eq4}
\end{gather}
}%

Similarly, $\mathsf{SF}_{\mathsf{EOD}}$ is defined using the following equation. In this case, only the true-positive changes are considered as suggested by the definition of EOD from \eqref{eq1}.

{\footnotesize
\begin{gather}
    \mathsf{SFC}_{i{\mathsf{EOD}}} =
    \begin{cases}
        1 & \text{if } Y_i = \hat{Y}_i(\mathcal{P}) = 1 \text{ and } \hat{Y}_i(\mathcal{P^*}) = 0 \nonumber \\
        -1 & \text{if } \hat{Y}_i(\mathcal{P}) = 0 \text{ and } Y_i = \hat{Y}_i(\mathcal{P^*}) = 1 \nonumber \\
        0 & \text{otherwise}
    \end{cases}
    \\
    \mathsf{SFC}_{\mathsf{EOD}} = \sum_{i=1}^{n} \mathsf{SFC}_{i{\mathsf{EOD}}} \nonumber
    \\
    \mathsf{SFR}_{\mathsf{EOD}}(u) = \mathsf{SFC}_{\mathsf{EOD}}(u)/n_{u, Y=1} \text{, } \mathsf{SFR}_{\mathsf{EOD}}(p) = \mathsf{SFC}_{\mathsf{EOD}}(p)/n_{p, Y=1} \nonumber 
    \\
    \mathsf{SF}_{\mathsf{EOD}} = \mathsf{SFR}_{\mathsf{EOD}}(u) - \mathsf{SFR}_{\mathsf{EOD}}(p)
\end{gather}
}%

Since AOD computes the average of true positive (TP) rate and false positive (FP) rate, first the change set for TP and FP predictions is computed. Then averaging the probability of changes for TP and FP, the change rates are computed for both groups. Finally, $\mathsf{SF}_{\mathsf{AOD}}$ is calculated by taking the difference of rates between privileged and unprivileged groups.

{\footnotesize
\begin{gather}
    \mathsf{SFC}_{i{\mathsf{TP}}} =
    \begin{cases}
        1 & \text{if } Y_i = \hat{Y}_i(\mathcal{P}) = 1 \text{ and } \hat{Y}_i(\mathcal{P^*}) = 0 \nonumber \\
        -1 & \text{if } \hat{Y}_i(\mathcal{P}) = 0 \text{ and } Y_i = \hat{Y}_i(\mathcal{P^*}) = 1 \nonumber \\
        0 & \text{otherwise}
    \end{cases}
    \\
    \mathsf{SFC}_{i{\mathsf{FP}}} =
    \begin{cases}
        1 & \text{if } \hat{Y}_i(\mathcal{P}) = 1 \text{ and } Y_i = \hat{Y}_i(\mathcal{P^*}) = 0 \nonumber \\
        -1 & \text{if } Y_i = \hat{Y}_i(\mathcal{P}) = 0 \text{ and } \hat{Y}_i(\mathcal{P^*}) = 1 \nonumber \\
        0 & \text{otherwise}
    \end{cases}
    \\
    \mathsf{SFC}_{\mathsf{TP}} = \sum_{i=1}^{n} \mathsf{SFC}_{i{\mathsf{TP}}} \text{ ,  } \mathsf{SFC}_{\mathsf{FP}} = \sum_{i=1}^{n} \mathsf{SFC}_{i{\mathsf{FP}}} \nonumber
    \\
    \mathsf{SFR}_{\mathsf{AOD}}(u) = (1/2) \{\mathsf{SFC}_{\mathsf{TP}}(u)/n_{u, Y=1} + \mathsf{SFC}_{\mathsf{FP}}(u)/n_{u, Y=0}\} \nonumber
    \\
    \mathsf{SFR}_{\mathsf{AOD}}(p) = (1/2) \{\mathsf{SFC}_{\mathsf{TP}}(p)/n_{p, Y=1} + \mathsf{SFC}_{\mathsf{FP}}(p)/n_{p, Y=0}\} \nonumber
    \\
    \mathsf{SF}_{\mathsf{AOD}} = \mathsf{SFR}_{\mathsf{AOD}}(u) - \mathsf{SFR}_{\mathsf{AOD}}(p)
\end{gather}
}%

Finally, $\mathsf{SF}_{\mathsf{ERD}}$ is computed using the change of count in both false positives (FP) and false negatives (FN) as mentioned in the definition of ERD in \eqref{eq1}.

{\footnotesize
\begin{gather}
    \mathsf{SFC}_{i{\mathsf{FN}}} =
    \begin{cases}
        1 & \text{if } \hat{Y}_i(\mathcal{P}) = 0 \text{ and } Y_i = \hat{Y}_i(\mathcal{P^*}) = 1 \nonumber \\
        -1 & \text{if } Y_i = \hat{Y}_i(\mathcal{P}) = 1 \text{ and } \hat{Y}_i(\mathcal{P^*}) = 0 \nonumber \\
        0 & \text{otherwise}
    \end{cases}
    \\
    \mathsf{SFC}_{\mathsf{FN}} = \sum_{i=1}^{n} \mathsf{SFC}_{i{\mathsf{FN}}} \nonumber
    \\
    \mathsf{SFR}_{\mathsf{ERR}}(u) = \mathsf{SFC}_{\mathsf{FP}}(u)/n_{u, Y=0} + \mathsf{SFC}_{\mathsf{FN}}(u)/n_{u, Y=1} \nonumber
    \\
    \mathsf{SFR}_{\mathsf{ERR}}(p) = \mathsf{SFC}_{\mathsf{FP}}(p)/n_{p, Y=0} + \mathsf{SFC}_{\mathsf{FN}}(p)/n_{p, Y=1} \nonumber
    \\
    \mathsf{SF}_{\mathsf{ERR}} = \mathsf{SFR}_{\mathsf{ERR}}(u) - \mathsf{SFR}_{\mathsf{ERR}}(p)
\end{gather}
}%

Thus far, we have four fairness metrics ($\mathsf{SF}_{\mathsf{SPD}}$, $\mathsf{SF}_{\mathsf{EOD}}$, $\mathsf{SF}_{\mathsf{AOD}}$, and $\mathsf{SF}_{\mathsf{ERD}}$) to measure the fairness of the stage.
In general, the rates computed by each metric ($\mathsf{SFR}$) follow the same range of the original metrics [-1, 1]. Therefore, the above metrics have a range [-2, 2].
Positive values indicate bias towards unprivileged group, negative values indicate bias towards privileged group, and values very close to 0 indicate fair preprocessing stage.

\section{Evaluation}
\label{sec:evaluation}

In this section, we describe the benchmark dataset and pipelines that we used for evaluation. Then we present the experiment design and results for answering the research questions.

\begin{figure*}[t]
	\centering
	\includegraphics[width=.78\textwidth]{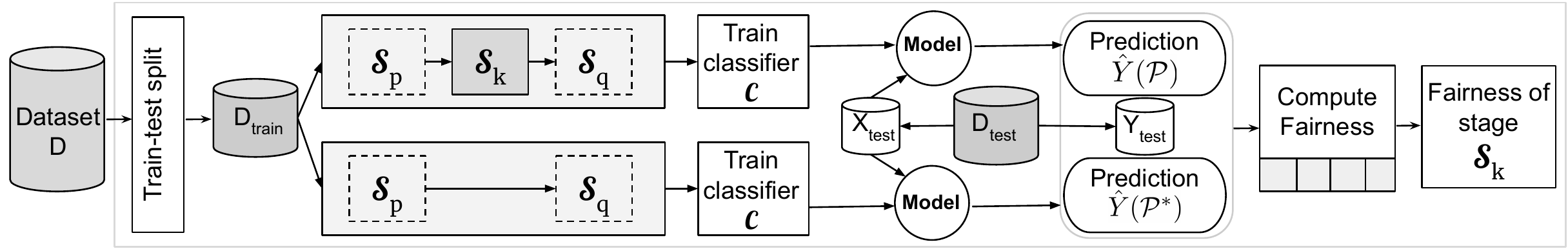}
	\caption{Experiment design to measure fairness of preprocessing stages in machine learning pipeline.}
	\label{experiment-design}
\end{figure*}

\subsection{Benchmark}
\label{subsec:benchmark}

We collected ML pipelines used in prior studies for fairness evaluation. First, Biswas and Rajan collected a benchmark of 40 ML models collected from Kaggle that operate on 5 different datasets e.g., \gc \cite{germanuci}, \ac \cite{kohavi1996scaling}, \bm \cite{moro2014data}, \hc \cite{home} and \ti \cite{titan}. 
However, the authors did not study the fairness at the component level, rather the ultimate fairness of the classifiers e.g., \texttt{RandomForest}, \texttt{DecisionTree}, etc. 
We revisited these Kaggle kernels and collected the preprocessing stages used in the pipelines. We noticed that \hc dataset \cite{home} in this benchmark is not unified like the other datasets, distributed over multiples CSV files, and the models under this dataset do not operate on the same data files. Hence, these models (8 out of 40) are not suitable for comparing fairness of data preprocessing stages.

Second, we collected the pipelines provided by \citeauthor{yang2020fairness} \cite{yang2020fairness}. The authors released 3 pipelines on two different datasets - \ac and \cp. 
Third, \citeauthor{zelaya2019towards} \cite{zelaya2019towards} studied the volatility of the preprocessing stages using two pipelines on a fairness dataset i.e., \gc. We included these pipelines in our benchmark.
Thus, we created a benchmark of 37 ML pipelines that operate on five datasets. 
The pipelines with the stages in each dataset category and their performances are shown in \tabref{tab:benchmark}.
Below we present a brief description of the datasets and associated tasks.

  \begin{table}[ht]
	\vspace{1.4mm}
	\footnotesize
	\centering
	\setlength\tabcolsep{1.5pt}
	\caption{The preprocessing stages and performance measures (accuracy, f1 score) of the pipelines in the benchmark}
	  \begin{tabular}{|l|r|c|c|l|r|c|c|}
	  \hline
	  \rowcolor[rgb]{ .251,  .251,  .251} \multicolumn{1}{|l|}{\textcolor[rgb]{ .949,  .949,  .949}{\textbf{German}}} & \multicolumn{1}{l|}{\textcolor[rgb]{ .949,  .949,  .949}{\textbf{Stages}}} & \multicolumn{1}{l|}{\textcolor[rgb]{ .949,  .949,  .949}{\textbf{Acc}}} & \multicolumn{1}{l|}{\textcolor[rgb]{ .949,  .949,  .949}{\textbf{F1}}} & \multicolumn{1}{l|}{\textcolor[rgb]{ .949,  .949,  .949}{\textbf{Adult}}} & \multicolumn{1}{l|}{\textcolor[rgb]{ .949,  .949,  .949}{\textbf{Stages}}} & \multicolumn{1}{l|}{\textcolor[rgb]{ .949,  .949,  .949}{\textbf{Acc}}} & \multicolumn{1}{l|}{\textcolor[rgb]{ .949,  .949,  .949}{\textbf{F1}}} \\
	  \hline
	  \multicolumn{1}{|l|}{GC1} & \multicolumn{1}{l|}{PCA, SB} & \multicolumn{1}{r|}{0.64} & \multicolumn{1}{r|}{0.76} & \multicolumn{1}{l|}{AC1} & \multicolumn{1}{l|}{SS, LE} & \multicolumn{1}{r|}{0.85} & \multicolumn{1}{r|}{0.66} \\
	  \hline
	  \rowcolor[rgb]{ .949,  .949,  .949} \multicolumn{1}{|l|}{GC2} & \multicolumn{1}{l|}{SMOTE, SS} & \multicolumn{1}{r|}{0.74} & \multicolumn{1}{r|}{0.81} & \multicolumn{1}{l|}{AC2} & \multicolumn{1}{l|}{MV} & \multicolumn{1}{r|}{\cellcolor[rgb]{ 1,  1,  1}0.85} & \multicolumn{1}{r|}{\cellcolor[rgb]{ 1,  1,  1}0.68} \\
	  \hline
	  \multicolumn{1}{|l|}{GC3} & \multicolumn{1}{l|}{PCA} & \multicolumn{1}{r|}{0.73} & \multicolumn{1}{r|}{0.83} & \multicolumn{1}{l|}{AC3} & \multicolumn{1}{l|}{Custom(f)} & \multicolumn{1}{r|}{0.87} & \multicolumn{1}{r|}{0.66} \\
	  \hline
	  \rowcolor[rgb]{ .949,  .949,  .949} \multicolumn{1}{|l|}{GC4} & \multicolumn{1}{l|}{LE, SS} & \multicolumn{1}{r|}{0.73} & \multicolumn{1}{r|}{0.82} & \multicolumn{1}{l|}{AC4} & \multicolumn{1}{l|}{PCA, SS} & \multicolumn{1}{r|}{\cellcolor[rgb]{ 1,  1,  1}0.85} & \multicolumn{1}{r|}{\cellcolor[rgb]{ 1,  1,  1}0.66} \\
	  \hline
	  \multicolumn{1}{|l|}{GC5} & \multicolumn{1}{l|}{SS} & \multicolumn{1}{r|}{0.74} & \multicolumn{1}{r|}{0.83} & \multicolumn{1}{l|}{AC5} & \multicolumn{1}{l|}{LE} & \multicolumn{1}{r|}{0.87} & \multicolumn{1}{r|}{0.71} \\
	  \hline
	  \rowcolor[rgb]{ .949,  .949,  .949} \multicolumn{1}{|l|}{GC6} & \multicolumn{1}{l|}{PCA, SS, LE} & \multicolumn{1}{r|}{0.73} & \multicolumn{1}{r|}{0.83} & \multicolumn{1}{l|}{AC6} & \multicolumn{1}{l|}{Custom(f), Custom(c)} & \multicolumn{1}{r|}{\cellcolor[rgb]{ 1,  1,  1}0.85} & \multicolumn{1}{r|}{\cellcolor[rgb]{ 1,  1,  1}0.65} \\
	  \hline
	  \multicolumn{1}{|l|}{GC7} & \multicolumn{1}{l|}{PCA, SB} & \multicolumn{1}{r|}{0.66} & \multicolumn{1}{r|}{0.77} & \multicolumn{1}{l|}{AC7} & \multicolumn{1}{l|}{PCA, SS, Custom(f)} & \multicolumn{1}{r|}{0.78} & \multicolumn{1}{r|}{0.51} \\
	  \hline
	  \rowcolor[rgb]{ .949,  .949,  .949} \multicolumn{1}{|l|}{GC8} & \multicolumn{1}{l|}{SS} & \multicolumn{1}{r|}{0.72} & \multicolumn{1}{r|}{0.81} & \multicolumn{1}{l|}{AC8} & \multicolumn{1}{l|}{SS, Custom(f), Stratify} & \multicolumn{1}{r|}{\cellcolor[rgb]{ 1,  1,  1}0.85} & \multicolumn{1}{r|}{\cellcolor[rgb]{ 1,  1,  1}0.67} \\
	  \hline
	  \multicolumn{1}{|l|}{GC9} & \multicolumn{1}{l|}{SMOTE} & \multicolumn{1}{r|}{0.67} & \multicolumn{1}{r|}{0.77} & \multicolumn{1}{l|}{AC9} & \multicolumn{1}{l|}{SS} & \multicolumn{1}{r|}{0.81} & \multicolumn{1}{r|}{0.61} \\
	  \hline
	  \rowcolor[rgb]{ .949,  .949,  .949} \multicolumn{1}{|l|}{GC10} & \multicolumn{1}{l|}{Usamp} & \multicolumn{1}{r|}{0.6} & \multicolumn{1}{r|}{0.81} & \multicolumn{1}{l|}{ACP10} & \multicolumn{1}{l|}{Impute} & \multicolumn{1}{r|}{\cellcolor[rgb]{ 1,  1,  1}0.81} & \multicolumn{1}{r|}{\cellcolor[rgb]{ 1,  1,  1}0.62} \\
	  \hline
	  \rowcolor[rgb]{ .251,  .251,  .251} \multicolumn{1}{|l|}{\textcolor[rgb]{ .949,  .949,  .949}{\textbf{Bank}}} & \multicolumn{1}{l|}{\textcolor[rgb]{ .949,  .949,  .949}{\textbf{Stages}}} & \multicolumn{1}{l|}{\textcolor[rgb]{ .949,  .949,  .949}{\textbf{Acc}}} & \multicolumn{1}{l|}{\textcolor[rgb]{ .949,  .949,  .949}{\textbf{F1}}} & \multicolumn{1}{l|}{\textcolor[rgb]{ .949,  .949,  .949}{\textbf{Titanic}}} & \multicolumn{1}{l|}{\textcolor[rgb]{ .949,  .949,  .949}{\textbf{Stages}}} & \multicolumn{1}{l|}{\textcolor[rgb]{ .949,  .949,  .949}{\textbf{Acc}}} & \multicolumn{1}{l|}{\textcolor[rgb]{ .949,  .949,  .949}{\textbf{F1}}} \\
	  \hline
	  \multicolumn{1}{|l|}{BM1} & \multicolumn{1}{l|}{Custom, LE, SS} & \multicolumn{1}{r|}{0.9} & \multicolumn{1}{r|}{0.56} & \multicolumn{1}{l|}{TT1} & \multicolumn{1}{l|}{MV, Custom(f), Encode} & \multicolumn{1}{r|}{0.77} & \multicolumn{1}{r|}{0.83} \\
	  \hline
	  \rowcolor[rgb]{ .949,  .949,  .949} \multicolumn{1}{|l|}{BM2} & \multicolumn{1}{l|}{LE} & \multicolumn{1}{r|}{0.91} & \multicolumn{1}{r|}{0.61} & \multicolumn{1}{l|}{TT2} & \multicolumn{1}{l|}{MV, Custom(f)} & \multicolumn{1}{r|}{\cellcolor[rgb]{ 1,  1,  1}0.78} & \multicolumn{1}{r|}{\cellcolor[rgb]{ 1,  1,  1}0.72} \\
	  \hline
	  \multicolumn{1}{|l|}{BM3} & \multicolumn{1}{l|}{LE, SS, Custom(f)} & \multicolumn{1}{r|}{0.9} & \multicolumn{1}{r|}{0.48} & \multicolumn{1}{l|}{TT3} & \multicolumn{1}{l|}{Custom(f), Impute} & \multicolumn{1}{r|}{0.8} & \multicolumn{1}{r|}{0.72} \\
	  \hline
	  \rowcolor[rgb]{ .949,  .949,  .949} \multicolumn{1}{|l|}{BM4} & \multicolumn{1}{l|}{SS} & \multicolumn{1}{r|}{0.89} & \multicolumn{1}{r|}{0.33} & \multicolumn{1}{l|}{TT4} & \multicolumn{1}{l|}{Custom(f), Impute, RFECV} & \multicolumn{1}{r|}{\cellcolor[rgb]{ 1,  1,  1}0.81} & \multicolumn{1}{r|}{\cellcolor[rgb]{ 1,  1,  1}0.73} \\
	  \hline
	  \multicolumn{1}{|l|}{BM5} & \multicolumn{1}{l|}{SS} & \multicolumn{1}{r}{0.88} & \multicolumn{1}{r|}{0.23} & \multicolumn{1}{l|}{TT5} & \multicolumn{1}{l|}{Custom(f)} & \multicolumn{1}{r|}{0.83} & \multicolumn{1}{r|}{0.76} \\
	  \hline
	  \rowcolor[rgb]{ .949,  .949,  .949} \multicolumn{1}{|l|}{BM6} & \multicolumn{1}{l|}{FS, Stratify, SS} & \multicolumn{1}{r|}{\cellcolor[rgb]{ 1,  1,  1}0.91} & \multicolumn{1}{r|}{\cellcolor[rgb]{ 1,  1,  1}0.58} & \multicolumn{1}{l|}{TT6} & \multicolumn{1}{l|}{Custom(f)} & \multicolumn{1}{r|}{\cellcolor[rgb]{ 1,  1,  1}0.82} & \multicolumn{1}{r|}{\cellcolor[rgb]{ 1,  1,  1}0.74} \\
	  \hline
	  \multicolumn{1}{|l|}{BM7} & \multicolumn{1}{l|}{FS} & \multicolumn{1}{r|}{0.91} & \multicolumn{1}{r|}{0.6} & \multicolumn{1}{l|}{TT7} & \multicolumn{1}{l|}{SS, LE, Custom(f)} & \multicolumn{1}{r|}{0.82} & \multicolumn{1}{r|}{0.77} \\
	  \hline
	  \rowcolor[rgb]{ .949,  .949,  .949} \multicolumn{1}{|l|}{BM8} & \multicolumn{1}{l|}{Stratify} & \multicolumn{1}{r|}{0.9} & \multicolumn{1}{r|}{0.56} & \multicolumn{1}{l|}{TT8} & \multicolumn{1}{l|}{Custom(f)} & \multicolumn{1}{r|}{\cellcolor[rgb]{ 1,  1,  1}0.83} & \multicolumn{1}{r|}{\cellcolor[rgb]{ 1,  1,  1}0.76} \\
	  \hline
	  \rowcolor[rgb]{ .251,  .251,  .251} \multicolumn{1}{|l|}{\textcolor[rgb]{ .949,  .949,  .949}{\textbf{Compas}}} & \multicolumn{5}{l|}{\textcolor[rgb]{ .949,  .949,  .949}{\textbf{Stages}}} & \multicolumn{1}{l|}{\textcolor[rgb]{ .949,  .949,  .949}{\textbf{Acc}}} & \multicolumn{1}{l|}{\textcolor[rgb]{ .949,  .949,  .949}{\textbf{F1}}} \\
	  \hline
	  \rowcolor[rgb]{ .949,  .949,  .949} \multicolumn{1}{|l|}{CP1} & \multicolumn{5}{l|}{Filter, Impute1, Encode, Impute2, Kbins, Binarize} & \multicolumn{1}{r|}{0.97} & \multicolumn{1}{r|}{0.97} \\
	  \hline
	  \multicolumn{8}{|p{8.3cm}|}{SB: SelectBest, SS: StandardScaler, LE*: LabelEncoder, Usamp: Undersampling, Custom(f/c): Custom \underline{f}eature engineering or \underline{c}leaning, FS: Feature selection, MV: Missing value processing, RFECV: Feature selection method} \\
	  \hline
	  \multicolumn{8}{wc{8.5cm}}{* Fairness is measured with respect to a reference stage.}
	  \end{tabular}%
	\label{tab:benchmark}%
  \end{table}%

\textit{German Credit.} The dataset contains 1000 data instances and 20 features of individuals who take credit from a bank \cite{germanuci}. The target is to classify whether the person has a good/bad credit risk.

\textit{Adult Census.} The dataset is extracted by Becker \cite{kohavi1996scaling} from 1994 census of United States. It contains 32,561 data instances and 12 features including demographic data of individuals. The task is to predict whether the person earns over 50K in a year. 

\textit{Bank Marketing.}
This dataset contains a bank's marketing campaign data of 41,188 individuals with 20 features \cite{moro2014data}. The goal is to classify whether a client will subscribe to a term deposit. 

\textit{Titanic.}
The dataset contains information about 891 passengers of Titanic \cite{titan}. The task is to predict the survival of the individuals on Titanic. The sensitive attribute of this dataset is \textit{sex}.

\textit{Compas.}
The dataset contains data of 6,889 criminal defendants in Florida. Propublica used this dataset and showed that the recidivism prediction software used in US courts discriminates between White and non-White \cite{angwin2016machine}. The task is to classify whether the defendants will re-offend where \textit{race} is considered as the sensitive attribute.

\subsection{Experiment Design}
\label{subsec:experiment}

Each pipeline in the benchmark consists of one or more preprocessing stages followed by the classifier.
In this paper, our main goal is to evaluate the fairness of different preprocessing stages using the fairness metrics described in \secref{sec:methodology}. 
The benchmark, code and results are released in the replication package \cite{biswas2021replication}.

The experiment design for evaluating the pipelines is shown in \fignref{experiment-design}. First, for each pipeline, we identified the preprocessing stages. For example, the pipeline in \secref{subsec:motivate1} contains six preprocessing stages. To evaluate the fairness of a stage $S_k$ in a pipeline $\mathcal{P}$, we create an alternative pipeline $\mathcal{P^*}$ by removing the stage $S_k$. For stages that can not be removed, we replaced $S_k$ with a reference stage $S_k'$. Among the preprocessing stages shown in \tabref{tab:benchmark}, we found only the encoders can not be removed. We experimented with all the encoders in Scikit-Learn library \cite{prepsklearn}, i.e., \texttt{OneHotEncoder}, \texttt{LabelEncoder}, \texttt{OrdinalEncoder}, and found that \texttt{OneHotEncoder} does not exhibit any bias. Therefore, we used \texttt{OneHotEncoder} as the reference stage for the encoders in our experiment.

Second, the original dataset is split into training (70\%) and test set (30\%). Then two copies of training data are used to train pipeline $\mathcal{P}$ and $\mathcal{P^*}$. After training the classifiers, two models predict the label for the same set of test data instances. Then, similar to the experimentation of \cite{zelaya2019towards}, for each prediction label we compare the two predictions $\hat{Y}_i(\mathcal{P})$ and $\hat{Y}_i(\mathcal{P}^*)$ with the true prediction label $Y_i$. This comparison provides the necessary data to compute the four fairness metrics. Similar to \cite{friedler2019comparative,biswas20machine}, for each stage in a pipeline, we run this experiment ten times, and then report the mean and standard deviation of the metrics, to avoid inconsistency of the randomness in the ML classifiers.
Finally, we followed the ML best practices so that noise is not introduced evaluating the fairness of preprocessing stages. For example, while applying some transformation, lack of data isolation might introduce noise in the evaluation, e.g., when applying \texttt{PCA} on dataset, it is important to train the \texttt{PCA} only using the training data. If we use the whole dataset to train the \texttt{PCA} and transform data, then information from the test-set might leak.
Third, since a stage operates on the data processed by the preceding stage(s), there are interdependencies between them. We always maintained the order of the stages while removing or replacing a stage (\secref{sec:stages}). To observe fairness of data transformers without interdependencies, we applied them on vanilla pipelines (\secref{sec:transformers}).

\section{Fairness of Preprocessing Stages}
\label{sec:stages}

\begin{figure*}[]
	\centering
	\includegraphics[width=\textwidth]{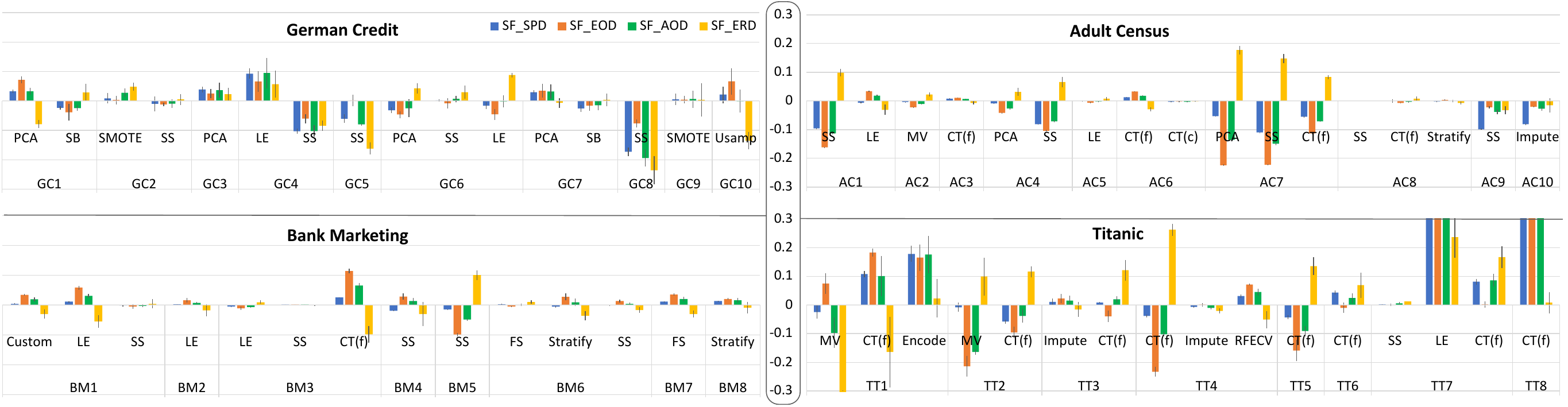}
	\caption{Fairness measures (Y-axis) of preprocessing stages in pipelines (X-axis). Grey lines above bars indicate standard error.}
	\label{experiment}
\end{figure*}

In this paper, we used a diverse set of metrics, to evaluate fairness of preprocessing stages. While developing an ML pipelines, if the developer has a comprehensive idea of the fairness of preprocessing stages, it would be convenient to build a fair pipeline. 
The evaluation has been done on 69 preprocessing stages in 37 ML pipelines from 5 dataset categories. For \cp dataset, we found one pipeline (\secref{subsec:motivate1}). Five out of six stages in this pipeline exhibit no bias, which has been discussed later.
For other 4 dataset categories, the evaluation has been shown in \fignref{experiment}. 
In this section, first, we discuss how we can interpret the metrics. Second, we answer the first research question and discuss the findings from our evaluation.

\subsection{What Do the Metrics Imply?}

We investigated the fairness of the preprocessing stages using four metrics: $\mathsf{SF}_{\mathsf{SPD}}$, $\mathsf{SF}_{\mathsf{EOD}}$, $\mathsf{SF}_{\mathsf{AOD}}$, and $\mathsf{SF}_{\mathsf{ERD}}$.  
These metrics measure the fairness of the stages by using the existing fairness criteria, e.g., $\mathsf{SF}_{\mathsf{SPD}}$ measures the fairness of a stage with respect to statistical parity difference ($\mathsf{SPD}$) criteria.
These fairness criteria evaluate algorithmic fairness of ML pipelines \cite{feldman2015certifying, zafar2015fairness, calders2010three}. The unfairness characterized by these criteria is measured based on the prediction disparities, although the root cause can be the training data or the algorithm (e.g., data preprocessing, classifier) itself. Therefore, when an ML model is identified as unfair, it implies that in the given predictive scenario, the outcome is biased. Similarly, the metrics proposed in this paper measure algorithmic unfairness caused by a specific preprocessing stage with respect to its pipeline.
For instance, in \fignref{experiment}, pipeline GC4 has two stages: \texttt{LabelEncoder} and \texttt{StandardScaler}.  
The fairness metrics suggest that \texttt{LabelEncoder} is biased towards unprivileged group (positive value), and \texttt{StandardScaler} is biased towards privileged group (negative value). The stages for which the measures are very close to zero, can be considered as fair preprocessing. 

The metrics can provide different fairness signals for a certain stage. For example, in AC4, $\mathsf{SF}_{\mathsf{ERD}}$ shows positive fairness, whereas the other metrics suggest negative fairness for both the stages \texttt{PCA} and \texttt{StandardScaler}. This disparity occurs because different metrics accounts for different fairness criteria. In this case, $\mathsf{SF}_{\mathsf{ERD}}$, depends on the false positive and false negative rate difference. No other metric is concerned about the false negative rate difference, and hence $\mathsf{SF}_{\mathsf{ERD}}$ provides a different fairness signal than other metrics. In practice, appropriate fairness criteria can vary depending on the task, usage scenario, or involved stakeholders. Study suggests that developers need to be aware of different fairness indicators to build fairer pipelines \cite{friedler2019comparative}. Therefore, we defined and evaluated fairness of stages with respect to multiple metrics.

\subsection{Fairness Analysis of Stages} 
The pipelines used both built-in algorithm imported from libraries i.e., \textit{data transformers} \cite{buitinck2013api}, as well as \textit{custom} preprocessing stages.
The stages found in each pipeline are shown in \tabref{tab:benchmark}, and the fairness measures of those stages are plotted in \fignref{experiment}.
Although the unfairness exhibited by a stage is with respect to the pipeline, we found fairness patterns of some stages and investigated them further.
In general, 
our findings show that the stages which change the underlying data distribution significantly, or modify minority data are responsible for increasing bias in the pipelines.

\finding{Data filtering and missing value removal change the data distribution and hence introduce bias in ML pipeline.}

Most of the real-world datasets contain missing values (MV) for several reasons such as data creation errors, not-applicable (N/A) attributes, incomplete data collection, etc. In our benchmark, \ac and \ti contain MV that required further processing in the pipeline. 7.4\% rows in \ac and 20.2\% rows in \ti have at least one missing feature in the dataset. The pipelines either remove the rows with MV or apply certain imputation \cite{imputer} technique that replaces the MV with mean, median or most frequently occurred values. Removal of rows with MV can significantly change the data distribution, which introduces bias in the pipeline. For example, both TT1 and TT2 removed data items with MV by applying \texttt{df.dropna()} method, which introduces bias in the prediction (\fignref{experiment}). Research has shown that MV are not uniformly distributed over all groups and data items from minority groups often contain more MV~\cite{dixon2018measuring}. If those data items are entirely removed, the representation of minority groups in the dataset becomes scarce.
On the other hand, TT3 applied mean-imputation and TT4 applied both median- and mode-imputation using \texttt{df.fillna()}, which exhibits fairness compared to data removal. 
While our findings suggest that removing data items with MV introduces bias, the most popular fairness tools AIF 360 \cite{bellamy2018ai}, Aequitas \cite{saleiro2018aequitas}, Themis-ML \cite{bantilan2018themis} ignore these data items and remove entire row/column. Our evaluation strategy confirms that the tools can integrate existing imputation methods \cite{imputer} in the pipeline and allow users to choose appropriate ones. Additionally, more research is needed to understand and develop imputation techniques that are fairness aware.

\finding{New feature generation or feature transformation can have large impact on fairness.}

We found that most of the feature engineering stages, especially the custom transformations exhibit bias in the pipeline.
For example, the pipelines in \ti dataset used custom feature engineering, since the dataset contains composite features which may provide additional information about the individuals. For instance, TT8 operates on the feature \textit{name} to create a new feature \textit{title} e.g., Mr, Mrs, Dr, etc. This transformation aims at better prediction of the survival of passengers by extracting the social status, but creates high bias between male and female.

In \ac, the feature \textit{eduction} of individuals contain values such as \textit{preschool}, \textit{10th}, \textit{1st-4th}, \textit{prof-school}, etc., which have been replaced by broad categories such \textit{Dropout}, \textit{HighGrad}, \textit{Masters} in the pipeline AC7. In addition, instead of using \textit{age} as continuous value, the feature has been discretized into $n$ number of bins. In both cases, the original data values have been modified, which has caused unfairness in the pipeline. Nevertheless, some pipelines (AC3, AC8) have custom feature transformations that are fair. 
Previous studies showed that certain features contribute more to the predictive quality of the model \cite{garreau2020explaining, ribeiro2018anchors}. Feature importance in prediction and corelation of features with the sensitive attribute also led to bias detection \cite{chakraborty2020making, grgic2018beyond} in ML models. However, does creating new features (by removing certain semantics) from a potentially biased feature increase the fairness, is an open question. Our method to quantify the fairness of such changes can guide further research in this direction.

\finding{Encoding techniques should be chosen cautiosly based on the classifier.}

Two most used encoding techniques for converting categorical feature to numerical feature are~\texttt{OneHotEncoder} and~\texttt{LabelEncoder}. \texttt{OneHotEncoder} creates $n$ new columns by replacing one column for each of the $n$ categories. \texttt{LabelEncoder} does not increase the number of the columns, and gives each category an integer label between $0$ and ($n-1$).
In our evaluation, we found that \texttt{LabelEncoder} introduces bias in \gc and \ti dataset but \texttt{OneHotEncoder} does not change fairness.
Since \texttt{LabelEncoder} imposes a sequential order between the categories, it might create a linear relation with the target value, and hence have an impact on the classifier to change fairness. For example, pipelines TT7 creates a new feature called \textit{Family} based on the surname of the person. This feature has a large number of unique categories (667 unique ones in 891 data instances). Therefore, the non-sparse representation in \textit{LabelEncoder} adds additional weight to the feature, which is causing unfairness in TT7. Developers might avoid \texttt{OneHotEncoder} because it suffers from the curse of dimensionality and the ordinal relation of data is lost. In that case, developers should be aware of the fairness impact of the encoder. One solution might be using PCA for dimensionality reduction, which has been done in GC7.

\finding{The variability of fairness of preprocessing stages depend on the dataset size and overall prediction rate of the pipelines.} 

We have plotted the standard error of the metrics as error bars in \fignref{experiment}. Firstly, it shows that the metrics in \gc and \ti dataset are more unstable. The reason is that the size of these two datasets is less than the other three datasets. \gc dataset has 1000 instances, and \ti has 891 instances. \ac and \bm dataset have more than 30K instances. If the sample size is large, data distribution tends to be similar even after taking a random train-test split \cite{friedler2019comparative}. However, when the dataset size is smaller, the distribution is changed among different train-test splitting. Furthermore, we have found that $\mathsf{SF}_{\mathsf{ERD}}$ is more unstable than other metrics. $\mathsf{SF}_{\mathsf{ERD}}$ depends on the change of false positive and false negative rates. However, in most cases, the pipelines are optimized for accuracy and precision, since these are some best performing ones collected from Kaggle. Therefore, before deploying preprocessing stages, it would be desirable to test the stability of over multiples executions. 

\begin{figure*}[!t]
	\centering
	\includegraphics[width=\textwidth]{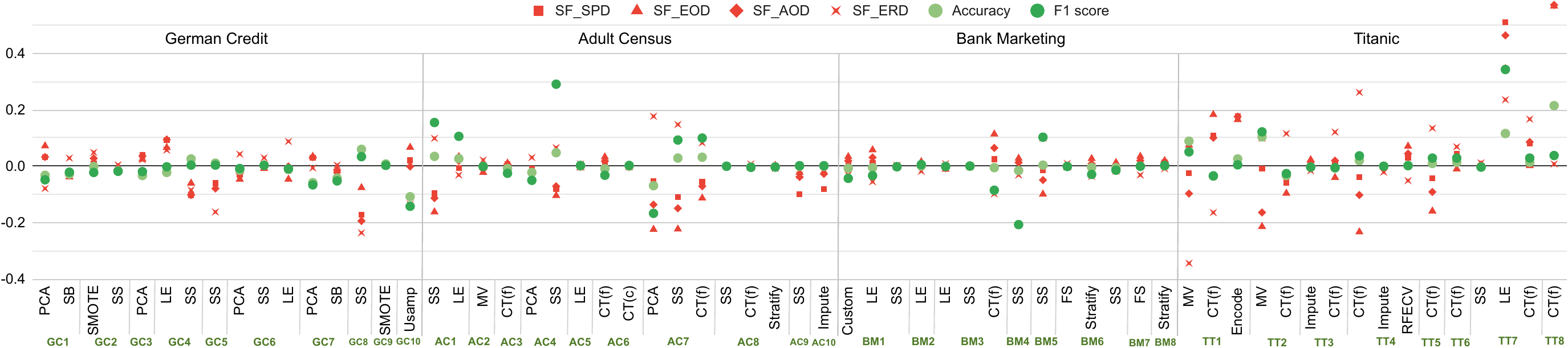}
	\caption{Performance changes (green) are plotted with fairness (red) of the preprocessing stages. A positive or negative performance change indicates performance increase or decrease respectively, after applying the stage.}
	\label{performance}
\end{figure*}

\finding{The unfairness of a preprocessing stage can be dominated by dataset or the classifier used in the pipeline.}

For the \cp dataset, we evaluated the six stages shown in \secref{subsec:motivate1}. All the stages exhibited data filtering show bias. The data filtering also showed bias close to zero (less than .005) with respect to all the metrics.
Although Yang \etal~\cite{yang2020fairness} argued that this pipeline filters data in different proportions from \textit{male} and \textit{female} group, our evaluation confirms it does not cause unfairness. This pipeline has been used by Propublica~\cite{angwin2016machine} to show the bias in the prediction. Therefore, it is understandable that they did not employ any preprocessing that introduces bias in the pipelines. 
Other than that, almost all the preprocessing stages in \bm pipelines also exhibit very little unfairness, which suggest that the preprocessing on this dataset are fair in general.

A few stages show different behavior when they are used in composition with different classifiers. For example, \texttt{StandardScaler} has been applied on both GC6 and GC8. While GC6 employs a \textit{RandomForest} classifier, GC8 uses \textit{K-Neighbors} classifier. We have observed the opposite fairness measures for \textit{StandardScaler} in these two pipelines. Therefore, fairness can be dominated by the underlying properties of data or the pipeline where it is applied. We have further investigated this phenomenon by applying transformers on different classifiers in the next section.

\subsection{Fairness-Performance Tradeoff}
In this section, we investigated the fairness-performance tradeoff for the preprocessing stages. The original performances of each pipeline have been reported in \tabref{tab:benchmark}. To investigate fairness of a stage, we created pipeline $\mathcal{P}^*$ by removing the stage from original pipeline $\mathcal{P}$ (\secref{experiment-design}). 
To understand fairness-performance tradeoff, we evaluated performance (both accuracy and f1 score) of $\mathcal{P}^*$ and $\mathcal{P}$ in the same experimental setup. Then we computed the performance difference to observe the impact of the stage on performance. For example, $Acc(\mathcal{P}) - Acc(\mathcal{P}^*)$ gives the accuracy increase (or decrease, if negative) after applying a stage. We plotted the performance impacts of the stages with their fairness measures in \fignref{performance}.

First, many preprocessing stages have negligible performance impact. In \fignref{performance}, 19 out 63 stages exhibits accuracy and f1 score change in the range [-0.005, 0.005], which indicates performance change $\le$ 0.05\%. We found that in all of these cases, except AC9 and AC10, the preprocessing stages are fair with a very small degree of bias. 
Second, tradeoff between performance and fairness is observed for the stages which improve performance. 17 stages improve accuracy or f1 score more than 0.05\%, which further exhibits moderate to high degree of bias. Overall, the most biased stages - TT7(LE), TT8(CT), TT4(CT), TT1(MV), GC8(SS), are improving performance. 
This stage-specific tradeoff is aligned with the overall performance-fairness tradeoff discussed in prior work \cite{biswas20machine, friedler2019comparative, chakraborty2019software}, which can be compared quantitatively by the work of \citeauthor{hort2021fairea} \cite{hort2021fairea}.
Third, we found that some stages decrease the performance, either accuracy or f1 score. Surprisingly, most of these stages also exhibit high degree of bias. For instance, the most performance-decreasing stages - BM4(SS), AC7(PCA), GC10 (Undersampling), are showing more bias. Our fairness evaluation would facilitate developers to identify and remove such stages in the pipeline.
\section{Fair Data Transformers}
\label{sec:transformers}

\begin{figure*}[t]
	\centering
	\includegraphics[width=\textwidth]{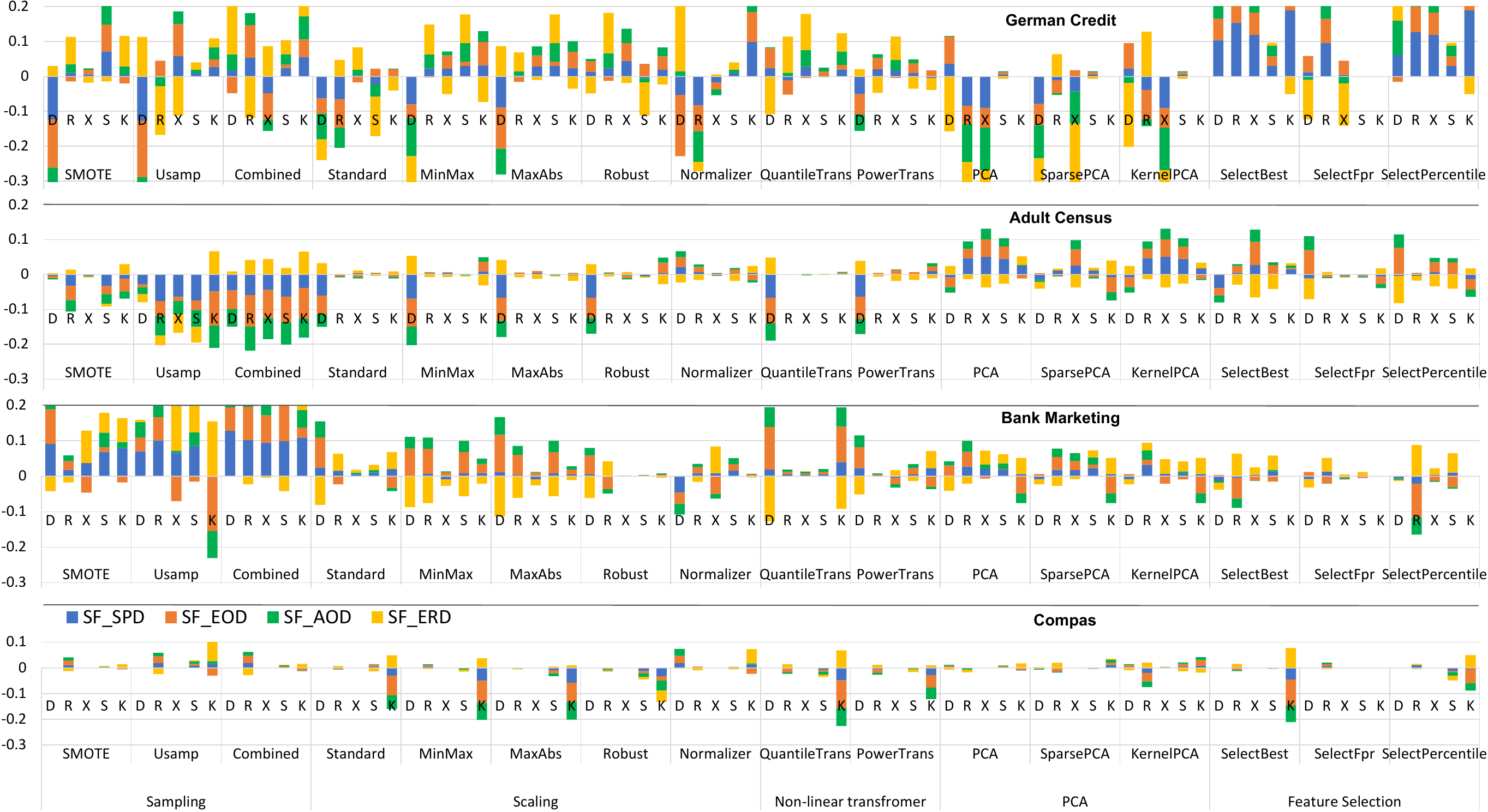}
	\caption{Fairness of transformers on classifiers, D: \texttt{DecisionTree}, R: \texttt{RandomForest}, X: \texttt{XGBoost}, S: \texttt{SupportVector}, K: \texttt{KNeighbors}}
	\label{rq2}
\end{figure*}

In \secref{sec:stages}, we found that many data preprocessing stages are biased. Many bias mitigation techniques applied in preprocessing stage have been shown successful~\cite{feldman2015certifying, kamiran2012data}.  
If we process data with appropriate transformer, then it might be possible to avoid bias and mitigate inherent bias in data or classifier. Even if a data transformer is biased towards a specific group, it could be useful to mitigate bias if original data or model exhibits bias towards the opposite group. To that end, we want to investigate the fairness pattern of the data transformers. However, in our evaluation (\fignref{experiment}), some transformers have been used only in specific situations e.g., \texttt{SMOTE} has been only applied on \gc dataset. What is the fairness of this transformer when used on other datasets and classifier? In this section, we setup experiments to evaluate the fairness of commonly used data transformers on different datasets and classifiers.

\begin{table}[]
    \vspace{4.5mm}
    \centering
    \footnotesize
    \setlength\tabcolsep{1pt}
    \caption{Transformers collected from pipelines and libraries}
      \begin{tabular}{|p{1.55cm}|p{2.6cm}|p{4.1cm}|}
      \hline
      \rowcolor[rgb]{ .251,  .251,  .251} \textcolor[rgb]{ .949,  .949,  .949}{\textbf{Categories}} & \textcolor[rgb]{ .949,  .949,  .949}{\textbf{Stages}} & \multicolumn{1}{l|}{\textcolor[rgb]{ .949,  .949,  .949}{\textbf{Transformers}}} \\
      MV processing & Imputation & SimpleImputer, IterativeImputer \\
      \hline
      Categorical encoding & \cellcolor[rgb]{ .851,  .851,  .851}Encoder & \cellcolor[rgb]{ .851,  .851,  .851}Binarizer, KBinsDiscretizer, LabelBinarizer, LabelEncoder, OneHotEncoder \\
      \hline
      \multirow{2}[2]{*}{Standardization} & Scaling & StandardScaler, MaxAbsScaler, MinMaxScaler, RobustScaler \\
            & \cellcolor[rgb]{ .851,  .851,  .851}Normalization & \cellcolor[rgb]{ .851,  .851,  .851}l1-normalizer, l2-normalizer \\
      \hline
      \multirow{3}[2]{*}{\makecell{Feature\\engineering}} & Non-linear transformation & QuantileTransformer, PowerTransformer \\
            & \cellcolor[rgb]{ .851,  .851,  .851}Polynomial feature generation & \cellcolor[rgb]{ .851,  .851,  .851}PCA, SparsePCA, MiniBatchSparsePCA, KernelPCA \\
            & Feature selection & SelectKBest, SelectFpr, SelectPercentile \\
      \hline
      \multirow{3}[2]{*}{Sampling} & \cellcolor[rgb]{ .851,  .851,  .851}Oversampling & \cellcolor[rgb]{ .851,  .851,  .851}SMOTE \\
            & Undersampling & AllKNN \\
            & \cellcolor[rgb]{ .851,  .851,  .851}Stratification & \cellcolor[rgb]{ .851,  .851,  .851}Random, Stratified \\
      \hline
      \end{tabular}%
    \label{tab:transformer}%
  \end{table}%

First, we collected the classifiers used in each dataset category from the benchmark. Then, for each dataset, we created a set of vanilla pipelines. A vanilla pipeline is a classification pipeline which contain only one classifier.
Second, we found a few categories of preprocessing stages from our benchmark shown in~\tabref{tab:transformer}. For each transformer used in each stage, we collected the alternative transformers from corresponding library.  
For example, in our benchmark, \texttt{StandardScaler} from Scikit-Learn library has been used for scaling data distribution in many pipelines. We collected other standardizing algorithms available in Scikit-Learn. We found that besides \texttt{StandardScaler}, Scikit-Learn also provides \texttt{MaxAbsScaler}, \texttt{MinMaxScaler}, and \texttt{Normalizer} standardize data \cite{prepsklearn}. Similarly, a data oversampling technique \texttt{SMOTE} has been used in the benchmark, we collected another undersampling technique \texttt{ALLKNN} and a combination of over- and undersampling sampling technique \texttt{SMOTENN} from IMBLearn library \cite{lemaitre2017imbalanced}.
Third, in each of the vanilla pipelines, we applied the transformers and evaluated fairness using the method used in \secref{subsec:experiment} with respect to four metrics.
We found that pipelines under \ti uses custom transformation, and most of the built-in transformers are not appropriate for this dataset. So, to be able to make the comparison consistent, we conducted this evaluation on four datasets: \gc, \ac, \bm, \cp. Finally, we did not use transformers for imputation and encoder stages. Encoding transformers (LabelEncoder, OneHotEncoder), have been applied on most of the pipelines and their behavior has been understood. The fairness measures of each transformer on different classifiers have been plotted in \fignref{rq2}. 

Fairness among the datasets follows a similar pattern. This further confirms that the unfairness is rooted in data. The \cp dataset shows the least bias. Although racial discrimination has been reported for this dataset~\cite{angwin2016machine}, this is a more curated dataset than the other three.
By looking at the overall trend of fairness, we observe that sampling techniques have the most biased impact on prediction. Other than that feature selection transformers have more impact than other ones.

\finding{Among all the transformers, applying sampling technique exhibits most unfairness.}

Sampling techniques are often used in ML tasks when dataset is class-imbalanced. 
Unlike the other transformers, sampling techniques make horizontal transformation to the training data. The oversampling technique \texttt{SMOTE} creates new data instances for the minority class by choosing the nearest data points in the feature space. Undersampling techniques balance dataset by removing data items from majority class. Although balancing dataset has been shown to increase fairness~\cite{dixon2018measuring}, our evaluation suggest that in three out of four datasets, it increases bias. 

From~\fignref{rq2}, we can see that sampling techniques exhibit the most unfairness. In \gc dataset, different classifier reacts differently when sampling is done. \texttt{DecisionTree} classifier exhibits most unfairness for both oversampling and undersampling towards privileged group i.e., \textit{male}.
Interestingly, the combination of over- and undersampling also fails to show fairness. 
Furthermore, both \gc and \bm pipelines exhibit bias towards unprivileged group, which might be desired when compared to bias towards privileged.

\finding{Selecting subset of features often increase unfairness.}

Selecting the best performing feature can give performance improvement of the pipeline. However, unfairness can be encoded in specific features~\cite{grgic2018beyond}. While selecting best features, some features which encodes unfairness, can dominate the outcome. Thus, many classifiers in \gc, \ac, and \bm show unfairness because of reduced number of features, which has been also observed by  
\citeauthor{zhang2021ignorance} \cite{zhang2018fairness}.
Surprisingly, \texttt{SelectFpr} exhibited very little or no bias compared to the other feature selection methods. A detailed investigation suggests that \texttt{SelectBest} and \texttt{SelectPercentile} select only the $k$ most contributing features. However, \texttt{SelectFpr} performs false positive rate test on each feature, and if it falls below a threshold, the feature is removed \cite{featselect}. Therefore, it does not apply harsh pruning, which contributes to the fairness of the prediction. 

\finding{In most of the pipelines, feature standardization and non-linear transformers are fair transformers.}

These transformers modify the mean and variance of the data by applying linear or non-linear transformation. However, they do not change the feature importance on the classifiers. Therefore, in most of the cases, these transformers (especially, \texttt{StandardScaler} and \texttt{RobustScaler}) are fair. Some classifiers show bias after applying these transformers such as, KNC in \cp. The unfairness exhibited by those pipelines are introduced by the classifiers, since these classifiers show similar bias pattern for other transformers as well.
The scalers can impact the fairness significantly if there are many outliers in data. That is why we see more bias for the scalers in \gc dataset. Therefore, although standardizing transformers are fair in general, they can be biased in composition with specific classifier or data property. 
\section{Fairness Composition of Stages}
\label{sec:composition}

From our evaluation, we found that many data transformers have fairness impact on ML pipeline.
In this section, we compare the \textit{local fairness} (fairness measures of preprocessing stages) with the \textit{global fairness} (fairness measures of whole pipeline).
First, we answer whether the local fairness composes in the global fairness. Second, we investigate if we can leverage the composition to mitigate bias by choosing appropriate transformers.

\subsection{Composition of Local and Global Fairness}

\begin{figure}[t]
	\centering
	\includegraphics[width=.99\columnwidth]{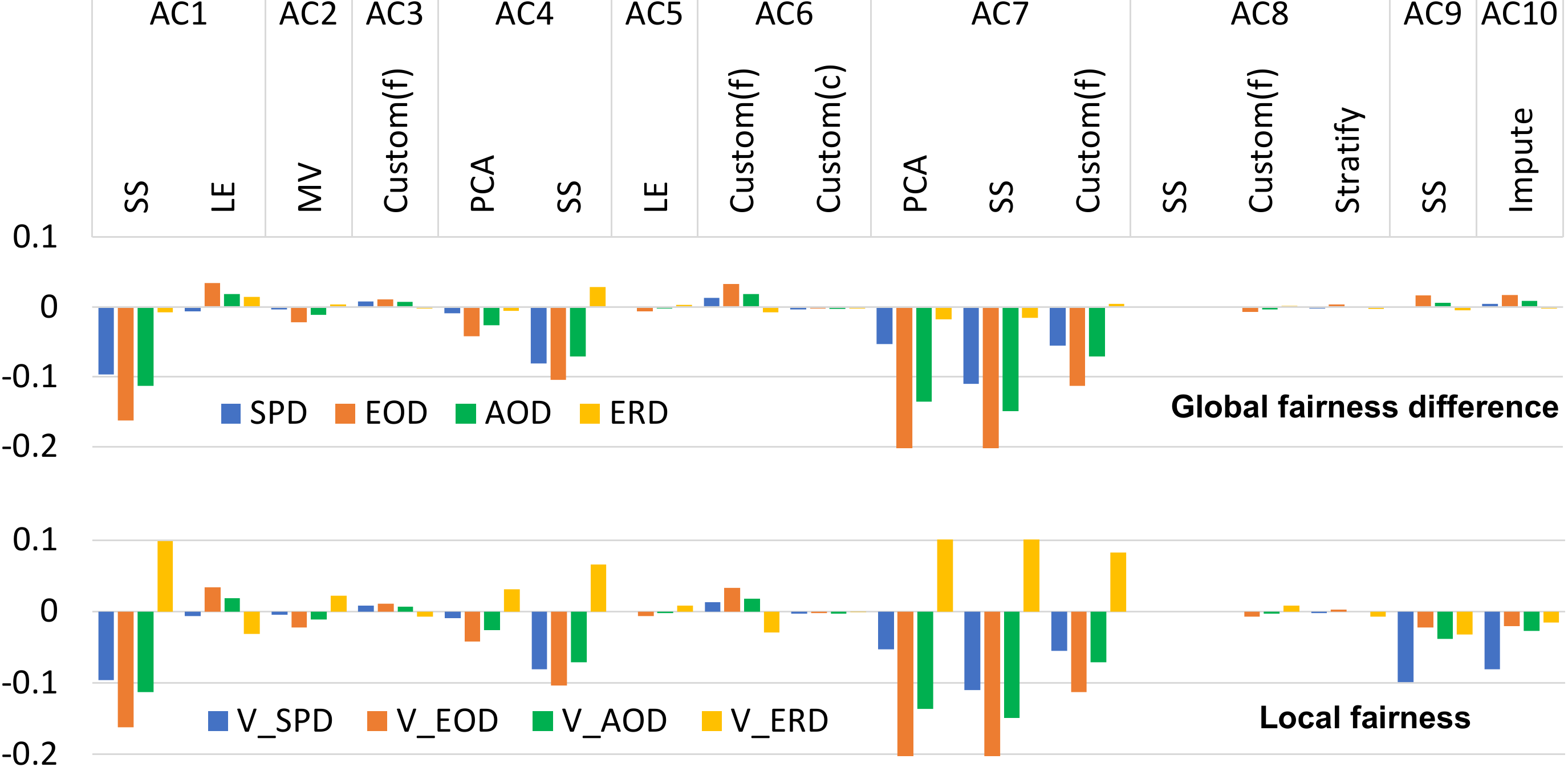}
	\caption{Comparison of global fairness change and local fairness for \ac dataset pipelines.}
	\label{local}
\end{figure}

We evaluated the global fairness of \ac pipelines (\tabref{tab:benchmark}) using the four existing metrics from \eqref{eq1}. We calculated the fairness difference of these pipelines before and after applying the preprocessing stages. Additionally, we have evaluated the stage-specific fairness metrics. Both the local fairness and difference in global fairness of those pipelines have been plotted in~\fignref{local}.

We can see that local and global fairness follow the same trend in most of the pipelines. This confirms that local fairness is directly contributing to the global fairness. 
However, the global fairness is computed based on the overall change in the prediction, whereas the local fairness considers the predictions for only those data instances which have been altered after applying a transformer \eqref{eq4}. For example, in \fignref{local}, for some pipelines (e.g., AC9, AC10), global and local fairness exhibit different trends. In these cases, the overall classification rate difference is not similar to the rate difference of altered labels. This means that the stages changed the labels such that it shows bias towards privileged. But when those changes in the labels are considered in addition to all the labels (global fairness), the bias difference could not capture the actual impact of that stage. 
We have verified this observation by manually inspecting the altered prediction labels. Thus, we can conclude that the local fairness composes to the global fairness. Specifically, if a preprocessing stage shows bias for privileged group, it pulls the global fairness towards the fairness direction of privileged group. However, only observing the global fairness difference, we can not measure the fairness of a given stage or transformer. 

\subsection{Bias Mitigation Using Appropriate Transformers}

For a given transformer in an ML pipeline, a \textit{downstream} transformer operates on data already processed by the given transformer and an upstream transformer is applied before the given one.
Since the fairness of a preprocessing stage composes to the global fairness, can we choose a downstream transformer to mitigate bias in ML pipeline? In this section, we empirically show that the global unfairness can be mitigated by choosing the appropriate downstream transformer.

\begin{figure}[h]
	\centering
	\includegraphics[width=.75\columnwidth]{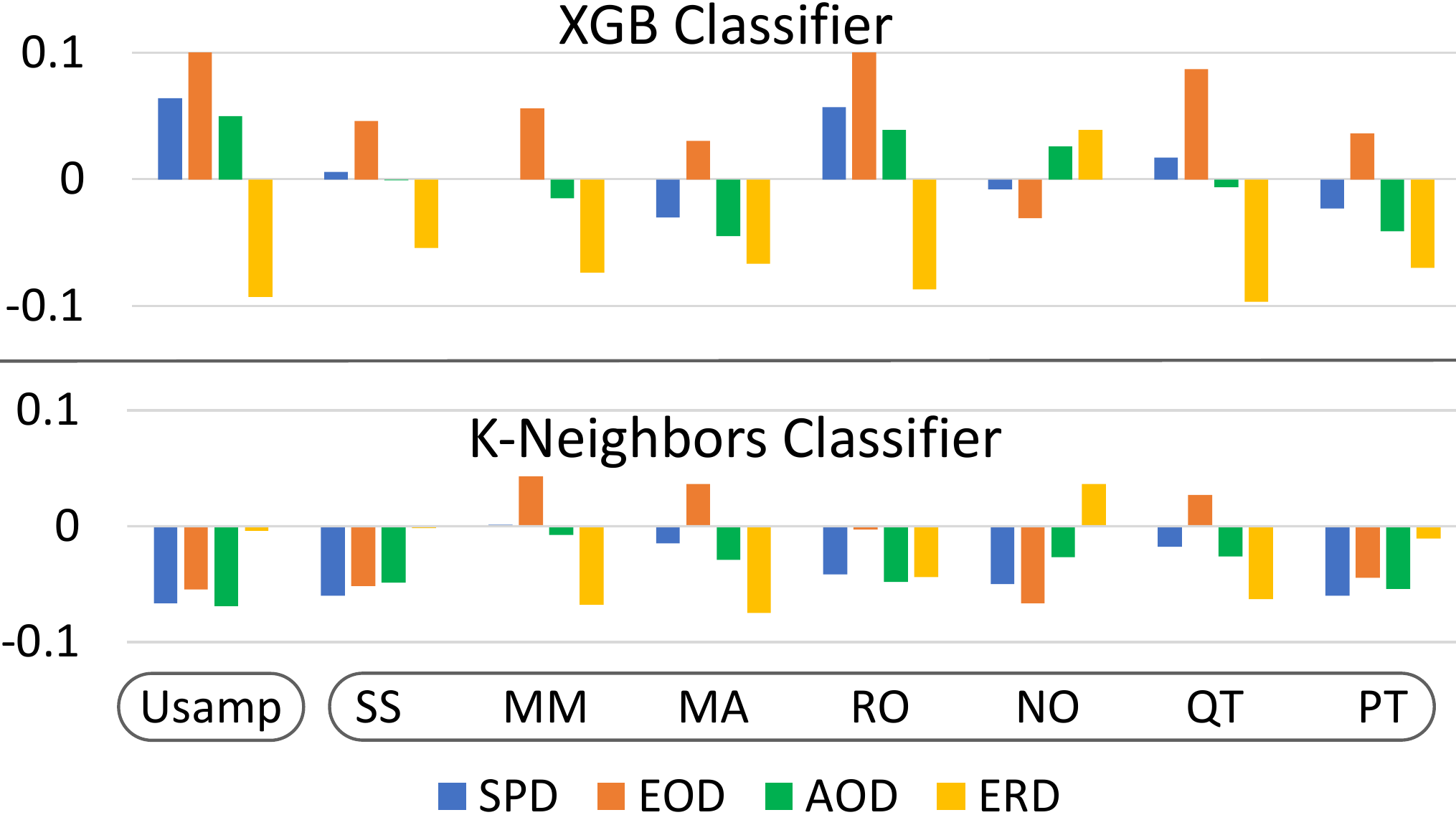}
	\caption{Global fairness after applying the upstream transformer (left), and after applying both the upstream and one downstream transformer (right). Usamp: undersampling.}
	\label{mitigation}
\end{figure}

Consider the use-case of classification task on \gc dataset with different classifiers similar to \fignref{rq2}. Suppose, the original pipeline is constructed using undersampling technique. Since this pipeline exhibits bias, as shown in \fignref{rq2}, can we choose a downstream data standardizing transformer that mitigates that bias? In this use case, undersampling is the upstream transformer, and any standardizing transformer is the downstream transformer. 

We showed the evaluation for XGB classifier and KNC classifier, since these two exhibits most bias when the upstream transformer was applied in \secnref{sec:transformers}.
We plotted the global fairness after applying only the upstream transformer in the left of \fignref{mitigation}.
We also reported the local fairness of the standardizing transformers in \tabref{tab:xgb-knc}. 
Now, since undersampling method exhibits bias towards privileged group for XGB, we look for the transformer that is biased towards privileged group. In \fignref{mitigation}, among other transformers, \texttt{Normalizer} is the most successful to mitigate bias of the upstream transformer.
Similarly, for KNC, the upstream operator exhibits bias towards privileged group. From~\tabref{tab:xgb-knc}, we can see that \texttt{MinMaxScaler} is the most biased transformer towards the opposite direction. As a result, applying \texttt{MinMaxScaler} mitigates bias the most. Note that the other downstream transformers also follow the fairness composition with its upstream transformer.
Therefore, by measuring fairness of the preprocessing stages, developers would be able to instrument the biased transformers and build fair ML pipelines.

\begin{table}[]
      \vspace{2mm}
      \centering
      \footnotesize
      \caption{Local fairness of stages as downstream transformer.}
      \label{tab:xgb-knc}
        \begin{tabular}{|l|l|wr{1.13cm}|wr{1.13cm}|wr{1.13cm}|wr{1.13cm}|}
        \hline
        \rowcolor[rgb]{ .251,  .251,  .251} \multicolumn{1}{|c}{\textcolor[rgb]{ .949,  .949,  .949}{\textbf{Model}}} & \textcolor[rgb]{ .949,  .949,  .949}{\textbf{Stage}} & \textcolor[rgb]{ .949,  .949,  .949}{\textbf{SF\_SPD}} & \textcolor[rgb]{ .949,  .949,  .949}{\textbf{SF\_EOD}} & \textcolor[rgb]{ .949,  .949,  .949}{\textbf{SF\_AOD}} & \textcolor[rgb]{ .949,  .949,  .949}{\textbf{SF\_ERD}} \\
        \multicolumn{1}{|c|}{\multirow{7}[1]{*}{XGB}} & SS    & \multicolumn{1}{r|}{0.001} & \multicolumn{1}{r|}{0.021} & \multicolumn{1}{r|}{-0.016} & \multicolumn{1}{r|}{-0.073} \\
              & \cellcolor[rgb]{ .851,  .851,  .851}MM & \multicolumn{1}{r|}{\cellcolor[rgb]{ .851,  .851,  .851}-0.006} & \multicolumn{1}{r|}{\cellcolor[rgb]{ .851,  .851,  .851}0.031} & \multicolumn{1}{r|}{\cellcolor[rgb]{ .851,  .851,  .851}-0.029} & \multicolumn{1}{r|}{\cellcolor[rgb]{ .851,  .851,  .851}-0.12} \\
              & MA    & \multicolumn{1}{r|}{-0.036} & \multicolumn{1}{r|}{0.005} & \multicolumn{1}{r|}{-0.059} & \multicolumn{1}{r|}{-0.128} \\
              & \cellcolor[rgb]{ .851,  .851,  .851}RO & \multicolumn{1}{r|}{\cellcolor[rgb]{ .851,  .851,  .851}0.051} & \multicolumn{1}{r|}{\cellcolor[rgb]{ .851,  .851,  .851}0.082} & \multicolumn{1}{r|}{\cellcolor[rgb]{ .851,  .851,  .851}0.025} & \multicolumn{1}{r|}{\cellcolor[rgb]{ .851,  .851,  .851}-0.113} \\
              & \cellcolor[rgb]{ 1,  .949,  .8}NO & \multicolumn{1}{r|}{\cellcolor[rgb]{ 1,  .949,  .8}-0.014} & \multicolumn{1}{r|}{\cellcolor[rgb]{ 1,  .949,  .8}-0.056} & \multicolumn{1}{r|}{\cellcolor[rgb]{ 1,  .949,  .8}0.012} & \multicolumn{1}{r|}{\cellcolor[rgb]{ 1,  .949,  .8}0.135} \\
              & \cellcolor[rgb]{ .851,  .851,  .851}QT & \multicolumn{1}{r|}{\cellcolor[rgb]{ .851,  .851,  .851}0.011} & \multicolumn{1}{r|}{\cellcolor[rgb]{ .851,  .851,  .851}0.062} & \multicolumn{1}{r|}{\cellcolor[rgb]{ .851,  .851,  .851}-0.02} & \multicolumn{1}{r|}{\cellcolor[rgb]{ .851,  .851,  .851}-0.165} \\
              & PT    & \multicolumn{1}{r|}{-0.029} & \multicolumn{1}{r|}{0.011} & \multicolumn{1}{r|}{-0.055} & \multicolumn{1}{r|}{-0.132} \\
        \hline
        \multicolumn{1}{|c|}{\multirow{7}[2]{*}{KNC}} & \cellcolor[rgb]{ .851,  .851,  .851}SS & \multicolumn{1}{r|}{\cellcolor[rgb]{ .851,  .851,  .851}0.035} & \multicolumn{1}{r|}{\cellcolor[rgb]{ .851,  .851,  .851}0.025} & \multicolumn{1}{r|}{\cellcolor[rgb]{ .851,  .851,  .851}0.055} & \multicolumn{1}{r|}{\cellcolor[rgb]{ .851,  .851,  .851}0.059} \\
              & \cellcolor[rgb]{ 1,  .949,  .8}MM & \multicolumn{1}{r|}{\cellcolor[rgb]{ 1,  .949,  .8}0.095} & \multicolumn{1}{r|}{\cellcolor[rgb]{ 1,  .949,  .8}0.12} & \multicolumn{1}{r|}{\cellcolor[rgb]{ 1,  .949,  .8}0.096} & \multicolumn{1}{r|}{\cellcolor[rgb]{ 1,  .949,  .8}-0.05} \\
              & \cellcolor[rgb]{ .851,  .851,  .851}MA & \multicolumn{1}{r|}{\cellcolor[rgb]{ .851,  .851,  .851}0.079} & \multicolumn{1}{r|}{\cellcolor[rgb]{ .851,  .851,  .851}0.114} & \multicolumn{1}{r|}{\cellcolor[rgb]{ .851,  .851,  .851}0.075} & \multicolumn{1}{r|}{\cellcolor[rgb]{ .851,  .851,  .851}-0.078} \\
              & RO    & \multicolumn{1}{r|}{0.052} & \multicolumn{1}{r|}{0.074} & \multicolumn{1}{r|}{0.056} & \multicolumn{1}{r|}{-0.036} \\
              & \cellcolor[rgb]{ .851,  .851,  .851}NO & \multicolumn{1}{r|}{\cellcolor[rgb]{ .851,  .851,  .851}0.045} & \multicolumn{1}{r|}{\cellcolor[rgb]{ .851,  .851,  .851}0.010} & \multicolumn{1}{r|}{\cellcolor[rgb]{ .851,  .851,  .851}0.077} & \multicolumn{1}{r|}{\cellcolor[rgb]{ .851,  .851,  .851}0.134} \\
              & QT    & \multicolumn{1}{r|}{0.077} & \multicolumn{1}{r|}{0.104} & \multicolumn{1}{r|}{0.078} & \multicolumn{1}{r|}{-0.052} \\
              & \cellcolor[rgb]{ .851,  .851,  .851}PT & \multicolumn{1}{r|}{\cellcolor[rgb]{ .851,  .851,  .851}0.035} & \multicolumn{1}{r|}{\cellcolor[rgb]{ .851,  .851,  .851}0.032} & \multicolumn{1}{r|}{\cellcolor[rgb]{ .851,  .851,  .851}0.050} & \multicolumn{1}{r|}{\cellcolor[rgb]{ .851,  .851,  .851}0.036} \\
              \hline \hline
              \multicolumn{6}{|p{7.6cm}|}{SS: StandardScaler, MM: MinMaxScaler, MA: MaxAbsScaler, RO: RobustScaler, NO: Normalizer, QT: QuantileTransformer, PT: PowerTransformer} \\
              \hline
              \end{tabular}%
              \vspace{-3mm}
        \end{table}%

\section{Discussion}
\label{sec:discussion}

We took the first step to understand the fairness of components in ML pipelines.
Our method helps to provide causality in software and reason about behavior of components based on the impact on outcome.
This method can be extended further to evaluate the fairness of other software modules \cite{pan20decomposing} in ML pipeline and localize faults \cite{wardat21deeplocalize}.
Moreover, we found most of the stages exhibited bias, to a low or higher degree.
The fairness measures of different components can be leveraged towards fairness-aware pipeline optimization to satisfy fairness constraints.
For example, US Equal Employment Commission suggests selection-rate difference between groups less than 20\% \cite{usemploy}.
Also, pipeline optimization techniques, e.g., TPOT \cite{olson2016tpot}, Lara \cite{kunft2019intermediate} can be potentially utilized for pipeline optimization.

Furthermore, research has been conducted to understand the impact of 
preprocessing stages with respect to performance improvement~\cite{crone2006impact, uysal2014impact, chandrasekar2016impact}. 
This paper will open research directions to develop preprocessing techniques that improve performance by keeping the fairness intact. 
We also reported a number of fairness patterns of preprocessing stages 
that inducing bias in the pipeline such as missing value processing, 
custom feature generation, feature selection. 
Moreover, instrumentation of the stages can mitigate the inherent bias of 
the classifiers. 
It shows opportunities to build automated tools for identifying fairness bugs in AI systems and recommending fixes \cite{islam20repairing, islam19}. Finally, current fairness tools (e.g., AIF 360 \cite{bellamy2018ai}, 
Aequitas \cite{saleiro2018aequitas}) can be augmented by incorporating data preprocessing stages into the pipelines and letting users have control over the data transformers and observe or mitigate bias. 
Similarly, the libraries can provide API support to monitor fairness of the transformers.
\section{Threats to Validity}
\label{sec:threats}

\textbf{Internal validity} refers to whether the fairness measures used in this paper actually captures the fairness of preprocessing stages. To mitigate this threat, we used existing concepts and metrics to build new set of metrics. 
Causality in software \cite{pearl2000causality, pearl2009causal} has been well-studied, and causal reasoning in fairness has also been popular \cite{kusner2017counterfactual, zhang2018fairness, salimi2019interventional, russell2017worlds}, since it can provide explanation with respect to change in the outcome. Besides, this method do not require an oracle because the prediction equivalences provide necessary information to measure the impact of the intervention \cite{galhotra2017fairness}.
Furthermore, in \secref{sec:composition}, we conducted experiments on local and global fairness to show how new metrics composes in the pipeline.

\textbf{External validity} is concerned about the extent the findings of this study can be generalized. To alleviate this threat, we conducted experiments on a large number of pipeline variations. We collected the pipelines from three different sources. Moreover, we collected alternative transformers from the ML libraries for comparative analysis. Finally, for the same dataset categories, we used multiple classifiers and fairness metrics so that the findings are persistent.
\section{Related Works}
\label{sec:related}

\paragraph{Fairness in ML Classification}
The machine learning community has defined different fairness criteria and proposed metrics to measure the fairness of classification tasks~\cite{zafar2015fairness, dixon2018measuring, pearl2009causal, dwork2012fairness, feldman2015certifying, hardt2016equality, calders2010three, chouldechova2017fair, zemel2013learning, speicher2018unified}. Following the measurement of fairness in ML models, many mitigation techniques have also been proposed to remove bias \cite{kamiran2012data, zhang2018mitigating, feldman2015certifying, dixon2018measuring, calders2010three, zafar2015fairness, chouldechova2017fair, goh2016satisfying, kamishima2012fairness, hardt2016equality, pleiss2017fairness, kamiran2012decision}. This body of work mostly concentrates on the theoretical aspect of fairness in a single classification task. 
Recently, software engineering community has also focused on the fairness in ML, mostly on fairness testing \cite{tramer2017fairtest, galhotra2017fairness, udeshi2018automated, aggarwal2019black}.
These works propose methods to generate appropriate test data inputs for the model and prediction on those inputs characterizes fairness.
Some research has been conducted to build automated tools \cite{adebayo2016iterative, udeshi2018automated, sokol2019fat} and libraries \cite{bellamy2018ai} for fairness. In addition, empirical studies have been conducted to compare, contrast between fairness aspects, interventions, tradeoffs, developers concerns, and human aspects of fairness \cite{friedler2019comparative,biswas20machine, harrison2020empirical, holstein2019improving, zhang2021ignorance}.

\textit{Fairness in Composition.}
Dwork and Ilvento argued that fairness is dynamic in a multi-component environment \cite{dwork2018fairness}.  
They showed that when multiple classifiers work in composition, even if the classifiers are fair in isolation, the overall system is not necessarily fair.
Bower \etal discussed fairness in ML \textit{pipeline}, where they considered \textit{pipeline} as sequence of multiple classification tasks \cite{bower2017fair}. They also showed that when decisions of fair components are compounded, the final decision might not be fair. For example, while interviewing candidates in two stages, fair decision in each stage may not guarantee a fair selection.  
D'Amour \etal studied the dynamics of fairness in multi-classification environnement using simulation \cite{d2020fairness}.
In these research, fairness composition is shown over multiple tasks and the authors did not consider fairness of components in single ML pipeline. We position our paper here to study the impact of preprocessing stages in ML pipeline and evaluate the fairness composition. 
\section{Conclusion}
\label{sec:conc}

Data preprocessing techniques are used in most of the machine learning pipelines in composition with the classifier. Studies showed that fairness of machine learning predictions depends largely on the data. In this paper, we investigated how the data preprocessing stages affect fairness of classification tasks. We proposed the causal method and leveraged existing metrics to measure the fairness of data preprocessing stages. The results showed that many stages induce bias in the prediction. By observing fairness of these data transformers, fairer ML pipelines can be built. In addition, we showed that existing bias can be mitigated by selecting appropriate transformers. 
We released the pipeline benchmark, code, and results to make our techniques available for further usages.
Future research can be conducted towards developing automated tools to detect bias in ML pipeline stages and instrument that accordingly.

\begin{acks}
	This work was supported in part by US NSF grants CNS-15-13263, CCF-19-34884, and Facebook Probability and Programming Award (809725759507969). We also thank the reviewers for their insightful comments. All opinions are of the authors and do not reflect the view of sponsors.
\end{acks}

\balance
\bibliographystyle{ACM-Reference-Format}
\bibliography{refs}


\begin{thebibliography}{77}


\ifx \showCODEN    \undefined \def \showCODEN     #1{\unskip}     \fi
\ifx \showDOI      \undefined \def \showDOI       #1{#1}\fi
\ifx \showISBNx    \undefined \def \showISBNx     #1{\unskip}     \fi
\ifx \showISBNxiii \undefined \def \showISBNxiii  #1{\unskip}     \fi
\ifx \showISSN     \undefined \def \showISSN      #1{\unskip}     \fi
\ifx \showLCCN     \undefined \def \showLCCN      #1{\unskip}     \fi
\ifx \shownote     \undefined \def \shownote      #1{#1}          \fi
\ifx \showarticletitle \undefined \def \showarticletitle #1{#1}   \fi
\ifx \showURL      \undefined \def \showURL       {\relax}        \fi
\providecommand\bibfield[2]{#2}
\providecommand\bibinfo[2]{#2}
\providecommand\natexlab[1]{#1}
\providecommand\showeprint[2][]{arXiv:#2}

\bibitem[\protect\citeauthoryear{Abadi, Barham, Chen, Chen, Davis, Dean, Devin,
  Ghemawat, Irving, Isard, et~al\mbox{.}}{Abadi et~al\mbox{.}}{2016}]%
        {abadi2016tensorflow}
\bibfield{author}{\bibinfo{person}{Mart{\'\i}n Abadi}, \bibinfo{person}{Paul
  Barham}, \bibinfo{person}{Jianmin Chen}, \bibinfo{person}{Zhifeng Chen},
  \bibinfo{person}{Andy Davis}, \bibinfo{person}{Jeffrey Dean},
  \bibinfo{person}{Matthieu Devin}, \bibinfo{person}{Sanjay Ghemawat},
  \bibinfo{person}{Geoffrey Irving}, \bibinfo{person}{Michael Isard},
  {et~al\mbox{.}}} \bibinfo{year}{2016}\natexlab{}.
\newblock \showarticletitle{Tensorflow: A system for large-scale machine
  learning}. In \bibinfo{booktitle}{\emph{12th {USENIX} symposium on operating
  systems design and implementation ({OSDI} 16)}}. \bibinfo{pages}{265--283}.
\newblock


\bibitem[\protect\citeauthoryear{Adebayo and Kagal}{Adebayo and Kagal}{2016}]%
        {adebayo2016iterative}
\bibfield{author}{\bibinfo{person}{Julius Adebayo} {and}
  \bibinfo{person}{Lalana Kagal}.} \bibinfo{year}{2016}\natexlab{}.
\newblock \showarticletitle{Iterative orthogonal feature projection for
  diagnosing bias in black-box models}.
\newblock \bibinfo{journal}{\emph{arXiv preprint arXiv:1611.04967}}
  (\bibinfo{year}{2016}).
\newblock


\bibitem[\protect\citeauthoryear{Aggarwal, Lohia, Nagar, Dey, and
  Saha}{Aggarwal et~al\mbox{.}}{2019}]%
        {aggarwal2019black}
\bibfield{author}{\bibinfo{person}{Aniya Aggarwal}, \bibinfo{person}{Pranay
  Lohia}, \bibinfo{person}{Seema Nagar}, \bibinfo{person}{Kuntal Dey}, {and}
  \bibinfo{person}{Diptikalyan Saha}.} \bibinfo{year}{2019}\natexlab{}.
\newblock \showarticletitle{Black box fairness testing of machine learning
  models}. In \bibinfo{booktitle}{\emph{Proceedings of the 2019 27th ACM Joint
  Meeting on European Software Engineering Conference and Symposium on the
  Foundations of Software Engineering}}. \bibinfo{pages}{625--635}.
\newblock


\bibitem[\protect\citeauthoryear{Amershi, Begel, Bird, DeLine, Gall, Kamar,
  Nagappan, Nushi, and Zimmermann}{Amershi et~al\mbox{.}}{2019}]%
        {amershi2019software}
\bibfield{author}{\bibinfo{person}{Saleema Amershi}, \bibinfo{person}{Andrew
  Begel}, \bibinfo{person}{Christian Bird}, \bibinfo{person}{Robert DeLine},
  \bibinfo{person}{Harald Gall}, \bibinfo{person}{Ece Kamar},
  \bibinfo{person}{Nachiappan Nagappan}, \bibinfo{person}{Besmira Nushi}, {and}
  \bibinfo{person}{Thomas Zimmermann}.} \bibinfo{year}{2019}\natexlab{}.
\newblock \showarticletitle{Software engineering for machine learning: A case
  study}. In \bibinfo{booktitle}{\emph{2019 IEEE/ACM 41st International
  Conference on Software Engineering: Software Engineering in Practice
  (ICSE-SEIP)}}. IEEE, \bibinfo{pages}{291--300}.
\newblock


\bibitem[\protect\citeauthoryear{Angwin, Larson, Mattu, and Kirchner}{Angwin
  et~al\mbox{.}}{2016}]%
        {angwin2016machine}
\bibfield{author}{\bibinfo{person}{Julia Angwin}, \bibinfo{person}{Jeff
  Larson}, \bibinfo{person}{Surya Mattu}, {and} \bibinfo{person}{Lauren
  Kirchner}.} \bibinfo{year}{2016}\natexlab{}.
\newblock \showarticletitle{Machine bias: There’s software used across the
  country to predict future criminals}.
\newblock \bibinfo{journal}{\emph{And it’s biased against blacks.
  ProPublica}} (\bibinfo{year}{2016}).
\newblock
\urldef\tempurl%
\url{https://github.com/propublica/compas-analysis}
\showURL{%
\tempurl}


\bibitem[\protect\citeauthoryear{Bantilan}{Bantilan}{2018}]%
        {bantilan2018themis}
\bibfield{author}{\bibinfo{person}{Niels Bantilan}.}
  \bibinfo{year}{2018}\natexlab{}.
\newblock \showarticletitle{Themis-ml: A fairness-aware machine learning
  interface for end-to-end discrimination discovery and mitigation}.
\newblock \bibinfo{journal}{\emph{Journal of Technology in Human Services}}
  \bibinfo{volume}{36}, \bibinfo{number}{1} (\bibinfo{year}{2018}),
  \bibinfo{pages}{15--30}.
\newblock


\bibitem[\protect\citeauthoryear{Baylor, Breck, Cheng, Fiedel, Foo, Haque,
  Haykal, Ispir, Jain, Koc, et~al\mbox{.}}{Baylor et~al\mbox{.}}{2017}]%
        {baylor2017tfx}
\bibfield{author}{\bibinfo{person}{Denis Baylor}, \bibinfo{person}{Eric Breck},
  \bibinfo{person}{Heng-Tze Cheng}, \bibinfo{person}{Noah Fiedel},
  \bibinfo{person}{Chuan~Yu Foo}, \bibinfo{person}{Zakaria Haque},
  \bibinfo{person}{Salem Haykal}, \bibinfo{person}{Mustafa Ispir},
  \bibinfo{person}{Vihan Jain}, \bibinfo{person}{Levent Koc}, {et~al\mbox{.}}}
  \bibinfo{year}{2017}\natexlab{}.
\newblock \showarticletitle{{TFX}: A tensorflow-based production-scale machine
  learning platform}. In \bibinfo{booktitle}{\emph{Proceedings of the 23rd ACM
  SIGKDD International Conference on Knowledge Discovery and Data Mining}}.
  \bibinfo{pages}{1387--1395}.
\newblock


\bibitem[\protect\citeauthoryear{Bellamy, Dey, Hind, Hoffman, Houde, Kannan,
  Lohia, Martino, Mehta, Mojsilovic, et~al\mbox{.}}{Bellamy
  et~al\mbox{.}}{2018}]%
        {bellamy2018ai}
\bibfield{author}{\bibinfo{person}{Rachel~KE Bellamy}, \bibinfo{person}{Kuntal
  Dey}, \bibinfo{person}{Michael Hind}, \bibinfo{person}{Samuel~C Hoffman},
  \bibinfo{person}{Stephanie Houde}, \bibinfo{person}{Kalapriya Kannan},
  \bibinfo{person}{Pranay Lohia}, \bibinfo{person}{Jacquelyn Martino},
  \bibinfo{person}{Sameep Mehta}, \bibinfo{person}{Aleksandra Mojsilovic},
  {et~al\mbox{.}}} \bibinfo{year}{2018}\natexlab{}.
\newblock \showarticletitle{{AI Fairness 360}: An extensible toolkit for
  detecting, understanding, and mitigating unwanted algorithmic bias}.
\newblock \bibinfo{journal}{\emph{arXiv preprint arXiv:1810.01943}}
  (\bibinfo{year}{2018}).
\newblock


\bibitem[\protect\citeauthoryear{Binns}{Binns}{2017}]%
        {binns2017fairness}
\bibfield{author}{\bibinfo{person}{Reuben Binns}.}
  \bibinfo{year}{2017}\natexlab{}.
\newblock \showarticletitle{Fairness in machine learning: Lessons from
  political philosophy}.
\newblock \bibinfo{journal}{\emph{arXiv preprint arXiv:1712.03586}}
  (\bibinfo{year}{2017}).
\newblock


\bibitem[\protect\citeauthoryear{Biswas and Rajan}{Biswas and Rajan}{2020}]%
        {biswas20machine}
\bibfield{author}{\bibinfo{person}{Sumon Biswas} {and} \bibinfo{person}{Hridesh
  Rajan}.} \bibinfo{year}{2020}\natexlab{}.
\newblock \showarticletitle{Do the Machine Learning Models on a Crowd Sourced
  Platform Exhibit Bias? An Empirical Study on Model Fairness}. In
  \bibinfo{booktitle}{\emph{Proceedings of the 28th ACM Joint Meeting on
  European Software Engineering Conference and Symposium on the Foundations of
  Software Engineering}} (Virtual Event, USA). \bibinfo{pages}{642–653}.
\newblock
\urldef\tempurl%
\url{https://doi.org/10.1145/3368089.3409704}
\showURL{%
\tempurl}


\bibitem[\protect\citeauthoryear{Biswas and Rajan}{Biswas and Rajan}{2021}]%
        {biswas2021replication}
\bibfield{author}{\bibinfo{person}{Sumon Biswas} {and} \bibinfo{person}{Hridesh
  Rajan}.} \bibinfo{year}{2021}\natexlab{}.
\newblock \showarticletitle{Replication Package for "Fair Preprocessing:
  Towards Understanding Compositional Fairness of Data Transformers in Machine
  Learning Pipeline"}.
\newblock  (\bibinfo{year}{2021}).
\newblock
\urldef\tempurl%
\url{https://github.com/sumonbis/FairPreprocessing}
\showURL{%
\tempurl}


\bibitem[\protect\citeauthoryear{Bower, Kitchen, Niss, Strauss, Vargas, and
  Venkatasubramanian}{Bower et~al\mbox{.}}{2017}]%
        {bower2017fair}
\bibfield{author}{\bibinfo{person}{Amanda Bower}, \bibinfo{person}{Sarah~N
  Kitchen}, \bibinfo{person}{Laura Niss}, \bibinfo{person}{Martin~J Strauss},
  \bibinfo{person}{Alexander Vargas}, {and} \bibinfo{person}{Suresh
  Venkatasubramanian}.} \bibinfo{year}{2017}\natexlab{}.
\newblock \showarticletitle{Fair pipelines}.
\newblock \bibinfo{journal}{\emph{arXiv preprint arXiv:1707.00391}}
  (\bibinfo{year}{2017}).
\newblock


\bibitem[\protect\citeauthoryear{Brun and Meliou}{Brun and Meliou}{2018}]%
        {brun2018software}
\bibfield{author}{\bibinfo{person}{Yuriy Brun} {and} \bibinfo{person}{Alexandra
  Meliou}.} \bibinfo{year}{2018}\natexlab{}.
\newblock \showarticletitle{Software fairness}. In
  \bibinfo{booktitle}{\emph{Proceedings of the 2018 26th ACM Joint Meeting on
  European Software Engineering Conference and Symposium on the Foundations of
  Software Engineering}}. \bibinfo{pages}{754--759}.
\newblock


\bibitem[\protect\citeauthoryear{Buitinck, Louppe, Blondel, Pedregosa, Mueller,
  Grisel, Niculae, Prettenhofer, Gramfort, Grobler, et~al\mbox{.}}{Buitinck
  et~al\mbox{.}}{2013}]%
        {buitinck2013api}
\bibfield{author}{\bibinfo{person}{Lars Buitinck}, \bibinfo{person}{Gilles
  Louppe}, \bibinfo{person}{Mathieu Blondel}, \bibinfo{person}{Fabian
  Pedregosa}, \bibinfo{person}{Andreas Mueller}, \bibinfo{person}{Olivier
  Grisel}, \bibinfo{person}{Vlad Niculae}, \bibinfo{person}{Peter
  Prettenhofer}, \bibinfo{person}{Alexandre Gramfort}, \bibinfo{person}{Jaques
  Grobler}, {et~al\mbox{.}}} \bibinfo{year}{2013}\natexlab{}.
\newblock \showarticletitle{{API} design for machine learning software:
  experiences from the scikit-learn project}.
\newblock \bibinfo{journal}{\emph{arXiv preprint arXiv:1309.0238}}
  (\bibinfo{year}{2013}).
\newblock


\bibitem[\protect\citeauthoryear{Calders and Verwer}{Calders and
  Verwer}{2010}]%
        {calders2010three}
\bibfield{author}{\bibinfo{person}{Toon Calders} {and} \bibinfo{person}{Sicco
  Verwer}.} \bibinfo{year}{2010}\natexlab{}.
\newblock \showarticletitle{Three naive Bayes approaches for
  discrimination-free classification}.
\newblock \bibinfo{journal}{\emph{Data Mining and Knowledge Discovery}}
  \bibinfo{volume}{21}, \bibinfo{number}{2} (\bibinfo{year}{2010}),
  \bibinfo{pages}{277--292}.
\newblock


\bibitem[\protect\citeauthoryear{Chakraborty, Peng, and Menzies}{Chakraborty
  et~al\mbox{.}}{2020}]%
        {chakraborty2020making}
\bibfield{author}{\bibinfo{person}{Joymallya Chakraborty},
  \bibinfo{person}{Kewen Peng}, {and} \bibinfo{person}{Tim Menzies}.}
  \bibinfo{year}{2020}\natexlab{}.
\newblock \showarticletitle{Making fair ML software using trustworthy
  explanation}. In \bibinfo{booktitle}{\emph{2020 35th IEEE/ACM International
  Conference on Automated Software Engineering (ASE)}}. IEEE,
  \bibinfo{pages}{1229--1233}.
\newblock


\bibitem[\protect\citeauthoryear{Chakraborty, Xia, Fahid, and
  Menzies}{Chakraborty et~al\mbox{.}}{2019}]%
        {chakraborty2019software}
\bibfield{author}{\bibinfo{person}{Joymallya Chakraborty},
  \bibinfo{person}{Tianpei Xia}, \bibinfo{person}{Fahmid~M Fahid}, {and}
  \bibinfo{person}{Tim Menzies}.} \bibinfo{year}{2019}\natexlab{}.
\newblock \showarticletitle{Software engineering for fairness: A case study
  with hyperparameter optimization}.
\newblock \bibinfo{journal}{\emph{arXiv preprint arXiv:1905.05786}}
  (\bibinfo{year}{2019}).
\newblock


\bibitem[\protect\citeauthoryear{Chandrasekar and Qian}{Chandrasekar and
  Qian}{2016}]%
        {chandrasekar2016impact}
\bibfield{author}{\bibinfo{person}{Priyanga Chandrasekar} {and}
  \bibinfo{person}{Kai Qian}.} \bibinfo{year}{2016}\natexlab{}.
\newblock \showarticletitle{The impact of data preprocessing on the performance
  of a naive bayes classifier}. In \bibinfo{booktitle}{\emph{2016 IEEE 40th
  Annual Computer Software and Applications Conference (COMPSAC)}},
  Vol.~\bibinfo{volume}{2}. IEEE, \bibinfo{pages}{618--619}.
\newblock


\bibitem[\protect\citeauthoryear{Chouldechova}{Chouldechova}{2017}]%
        {chouldechova2017fair}
\bibfield{author}{\bibinfo{person}{Alexandra Chouldechova}.}
  \bibinfo{year}{2017}\natexlab{}.
\newblock \showarticletitle{Fair prediction with disparate impact: A study of
  bias in recidivism prediction instruments}.
\newblock \bibinfo{journal}{\emph{Big data}} \bibinfo{volume}{5},
  \bibinfo{number}{2} (\bibinfo{year}{2017}), \bibinfo{pages}{153--163}.
\newblock


\bibitem[\protect\citeauthoryear{Crone, Lessmann, and Stahlbock}{Crone
  et~al\mbox{.}}{2006}]%
        {crone2006impact}
\bibfield{author}{\bibinfo{person}{Sven~F Crone}, \bibinfo{person}{Stefan
  Lessmann}, {and} \bibinfo{person}{Robert Stahlbock}.}
  \bibinfo{year}{2006}\natexlab{}.
\newblock \showarticletitle{The impact of preprocessing on data mining: An
  evaluation of classifier sensitivity in direct marketing}.
\newblock \bibinfo{journal}{\emph{European Journal of Operational Research}}
  \bibinfo{volume}{173}, \bibinfo{number}{3} (\bibinfo{year}{2006}),
  \bibinfo{pages}{781--800}.
\newblock


\bibitem[\protect\citeauthoryear{D'Amour, Srinivasan, Atwood, Baljekar,
  Sculley, and Halpern}{D'Amour et~al\mbox{.}}{2020}]%
        {d2020fairness}
\bibfield{author}{\bibinfo{person}{Alexander D'Amour}, \bibinfo{person}{Hansa
  Srinivasan}, \bibinfo{person}{James Atwood}, \bibinfo{person}{Pallavi
  Baljekar}, \bibinfo{person}{D Sculley}, {and} \bibinfo{person}{Yoni
  Halpern}.} \bibinfo{year}{2020}\natexlab{}.
\newblock \showarticletitle{Fairness is not static: deeper understanding of
  long term fairness via simulation studies}. In
  \bibinfo{booktitle}{\emph{Proceedings of the 2020 Conference on Fairness,
  Accountability, and Transparency}}. \bibinfo{pages}{525--534}.
\newblock


\bibitem[\protect\citeauthoryear{Dixon, Li, Sorensen, Thain, and
  Vasserman}{Dixon et~al\mbox{.}}{2018}]%
        {dixon2018measuring}
\bibfield{author}{\bibinfo{person}{Lucas Dixon}, \bibinfo{person}{John Li},
  \bibinfo{person}{Jeffrey Sorensen}, \bibinfo{person}{Nithum Thain}, {and}
  \bibinfo{person}{Lucy Vasserman}.} \bibinfo{year}{2018}\natexlab{}.
\newblock \showarticletitle{Measuring and mitigating unintended bias in text
  classification}. In \bibinfo{booktitle}{\emph{Proceedings of the 2018
  AAAI/ACM Conference on AI, Ethics, and Society}}. \bibinfo{pages}{67--73}.
\newblock


\bibitem[\protect\citeauthoryear{Dwork, Hardt, Pitassi, Reingold, and
  Zemel}{Dwork et~al\mbox{.}}{2012}]%
        {dwork2012fairness}
\bibfield{author}{\bibinfo{person}{Cynthia Dwork}, \bibinfo{person}{Moritz
  Hardt}, \bibinfo{person}{Toniann Pitassi}, \bibinfo{person}{Omer Reingold},
  {and} \bibinfo{person}{Richard Zemel}.} \bibinfo{year}{2012}\natexlab{}.
\newblock \showarticletitle{Fairness through awareness}. In
  \bibinfo{booktitle}{\emph{Proceedings of the 3rd innovations in theoretical
  computer science conference}}. \bibinfo{pages}{214--226}.
\newblock


\bibitem[\protect\citeauthoryear{Dwork and Ilvento}{Dwork and Ilvento}{2018}]%
        {dwork2018fairness}
\bibfield{author}{\bibinfo{person}{Cynthia Dwork} {and}
  \bibinfo{person}{Christina Ilvento}.} \bibinfo{year}{2018}\natexlab{}.
\newblock \showarticletitle{Fairness under composition}.
\newblock \bibinfo{journal}{\emph{arXiv preprint arXiv:1806.06122}}
  (\bibinfo{year}{2018}).
\newblock


\bibitem[\protect\citeauthoryear{Feldman, Friedler, Moeller, Scheidegger, and
  Venkatasubramanian}{Feldman et~al\mbox{.}}{2015}]%
        {feldman2015certifying}
\bibfield{author}{\bibinfo{person}{Michael Feldman}, \bibinfo{person}{Sorelle~A
  Friedler}, \bibinfo{person}{John Moeller}, \bibinfo{person}{Carlos
  Scheidegger}, {and} \bibinfo{person}{Suresh Venkatasubramanian}.}
  \bibinfo{year}{2015}\natexlab{}.
\newblock \showarticletitle{Certifying and removing disparate impact}. In
  \bibinfo{booktitle}{\emph{proceedings of the 21th ACM SIGKDD international
  conference on knowledge discovery and data mining}}.
  \bibinfo{pages}{259--268}.
\newblock


\bibitem[\protect\citeauthoryear{Friedler, Scheidegger, Venkatasubramanian,
  Choudhary, Hamilton, and Roth}{Friedler et~al\mbox{.}}{2019}]%
        {friedler2019comparative}
\bibfield{author}{\bibinfo{person}{Sorelle~A Friedler}, \bibinfo{person}{Carlos
  Scheidegger}, \bibinfo{person}{Suresh Venkatasubramanian},
  \bibinfo{person}{Sonam Choudhary}, \bibinfo{person}{Evan~P Hamilton}, {and}
  \bibinfo{person}{Derek Roth}.} \bibinfo{year}{2019}\natexlab{}.
\newblock \showarticletitle{A comparative study of fairness-enhancing
  interventions in machine learning}. In \bibinfo{booktitle}{\emph{Proceedings
  of the Conference on Fairness, Accountability, and Transparency}}.
  \bibinfo{pages}{329--338}.
\newblock


\bibitem[\protect\citeauthoryear{Galhotra, Brun, and Meliou}{Galhotra
  et~al\mbox{.}}{2017}]%
        {galhotra2017fairness}
\bibfield{author}{\bibinfo{person}{Sainyam Galhotra}, \bibinfo{person}{Yuriy
  Brun}, {and} \bibinfo{person}{Alexandra Meliou}.}
  \bibinfo{year}{2017}\natexlab{}.
\newblock \showarticletitle{Fairness testing: testing software for
  discrimination}. In \bibinfo{booktitle}{\emph{Proceedings of the 2017 11th
  Joint Meeting on Foundations of Software Engineering}}.
  \bibinfo{pages}{498--510}.
\newblock


\bibitem[\protect\citeauthoryear{Garreau and Luxburg}{Garreau and
  Luxburg}{2020}]%
        {garreau2020explaining}
\bibfield{author}{\bibinfo{person}{Damien Garreau} {and}
  \bibinfo{person}{Ulrike Luxburg}.} \bibinfo{year}{2020}\natexlab{}.
\newblock \showarticletitle{Explaining the explainer: A first theoretical
  analysis of LIME}. In \bibinfo{booktitle}{\emph{International Conference on
  Artificial Intelligence and Statistics}}. PMLR, \bibinfo{pages}{1287--1296}.
\newblock


\bibitem[\protect\citeauthoryear{Goh, Cotter, Gupta, and Friedlander}{Goh
  et~al\mbox{.}}{2016}]%
        {goh2016satisfying}
\bibfield{author}{\bibinfo{person}{Gabriel Goh}, \bibinfo{person}{Andrew
  Cotter}, \bibinfo{person}{Maya Gupta}, {and} \bibinfo{person}{Michael~P
  Friedlander}.} \bibinfo{year}{2016}\natexlab{}.
\newblock \showarticletitle{Satisfying real-world goals with dataset
  constraints}. In \bibinfo{booktitle}{\emph{Advances in Neural Information
  Processing Systems}}. \bibinfo{pages}{2415--2423}.
\newblock


\bibitem[\protect\citeauthoryear{Goodall}{Goodall}{2016}]%
        {goodall2016can}
\bibfield{author}{\bibinfo{person}{Noah~J Goodall}.}
  \bibinfo{year}{2016}\natexlab{}.
\newblock \showarticletitle{Can you program ethics into a self-driving car?}
\newblock \bibinfo{journal}{\emph{IEEE Spectrum}} \bibinfo{volume}{53},
  \bibinfo{number}{6} (\bibinfo{year}{2016}), \bibinfo{pages}{28--58}.
\newblock


\bibitem[\protect\citeauthoryear{Grgic-Hlaca, Zafar, Gummadi, and
  Weller}{Grgic-Hlaca et~al\mbox{.}}{2018}]%
        {grgic2018beyond}
\bibfield{author}{\bibinfo{person}{Nina Grgic-Hlaca},
  \bibinfo{person}{Muhammad~Bilal Zafar}, \bibinfo{person}{Krishna~P Gummadi},
  {and} \bibinfo{person}{Adrian Weller}.} \bibinfo{year}{2018}\natexlab{}.
\newblock \showarticletitle{Beyond Distributive Fairness in Algorithmic
  Decision Making: Feature Selection for Procedurally Fair Learning.}. In
  \bibinfo{booktitle}{\emph{AAAI}}. \bibinfo{pages}{51--60}.
\newblock


\bibitem[\protect\citeauthoryear{Hardt, Price, and Srebro}{Hardt
  et~al\mbox{.}}{2016}]%
        {hardt2016equality}
\bibfield{author}{\bibinfo{person}{Moritz Hardt}, \bibinfo{person}{Eric Price},
  {and} \bibinfo{person}{Nati Srebro}.} \bibinfo{year}{2016}\natexlab{}.
\newblock \showarticletitle{Equality of opportunity in supervised learning}. In
  \bibinfo{booktitle}{\emph{Advances in neural information processing
  systems}}. \bibinfo{pages}{3315--3323}.
\newblock


\bibitem[\protect\citeauthoryear{Harrison, Hanson, Jacinto, Ramirez, and
  Ur}{Harrison et~al\mbox{.}}{2020}]%
        {harrison2020empirical}
\bibfield{author}{\bibinfo{person}{Galen Harrison}, \bibinfo{person}{Julia
  Hanson}, \bibinfo{person}{Christine Jacinto}, \bibinfo{person}{Julio
  Ramirez}, {and} \bibinfo{person}{Blase Ur}.} \bibinfo{year}{2020}\natexlab{}.
\newblock \showarticletitle{An empirical study on the perceived fairness of
  realistic, imperfect machine learning models}. In
  \bibinfo{booktitle}{\emph{Proceedings of the 2020 Conference on Fairness,
  Accountability, and Transparency}}. \bibinfo{pages}{392--402}.
\newblock


\bibitem[\protect\citeauthoryear{Hofmann}{Hofmann}{1994}]%
        {germanuci}
\bibfield{author}{\bibinfo{person}{Dr.~Hans Hofmann}.}
  \bibinfo{year}{1994}\natexlab{}.
\newblock \bibinfo{title}{German Credit Dataset: {UCI} Machine Learning
  Repository}.
\newblock
\newblock
\urldef\tempurl%
\url{https://archive.ics.uci.edu/ml/datasets/Statlog+(German+Credit+Data)}
\showURL{%
\tempurl}


\bibitem[\protect\citeauthoryear{Holstein, Wortman~Vaughan, Daum{\'e}~III,
  Dudik, and Wallach}{Holstein et~al\mbox{.}}{2019}]%
        {holstein2019improving}
\bibfield{author}{\bibinfo{person}{Kenneth Holstein}, \bibinfo{person}{Jennifer
  Wortman~Vaughan}, \bibinfo{person}{Hal Daum{\'e}~III}, \bibinfo{person}{Miro
  Dudik}, {and} \bibinfo{person}{Hanna Wallach}.}
  \bibinfo{year}{2019}\natexlab{}.
\newblock \showarticletitle{Improving fairness in machine learning systems:
  What do industry practitioners need?}. In
  \bibinfo{booktitle}{\emph{Proceedings of the 2019 CHI Conference on Human
  Factors in Computing Systems}}. \bibinfo{pages}{1--16}.
\newblock


\bibitem[\protect\citeauthoryear{Hort, Zhang, Sarro, and Harman}{Hort
  et~al\mbox{.}}{2021}]%
        {hort2021fairea}
\bibfield{author}{\bibinfo{person}{Max Hort}, \bibinfo{person}{Jie~M Zhang},
  \bibinfo{person}{Federica Sarro}, {and} \bibinfo{person}{Mark Harman}.}
  \bibinfo{year}{2021}\natexlab{}.
\newblock \showarticletitle{Fairea: A Model Behaviour Mutation Approach to
  Benchmarking Bias Mitigation Methods}. In
  \bibinfo{booktitle}{\emph{Proceedings of the 29th ACM Joint Meeting on
  European Software Engineering Conference and Symposium on the Foundations of
  Software Engineering (to appear)}} (Athens, Greece).
\newblock


\bibitem[\protect\citeauthoryear{Islam, Nguyen, Pan, and Rajan}{Islam
  et~al\mbox{.}}{2019}]%
        {islam19}
\bibfield{author}{\bibinfo{person}{Md~Johirul Islam}, \bibinfo{person}{Giang
  Nguyen}, \bibinfo{person}{Rangeet Pan}, {and} \bibinfo{person}{Hridesh
  Rajan}.} \bibinfo{year}{2019}\natexlab{}.
\newblock \showarticletitle{A Comprehensive Study on Deep Learning Bug
  Characteristics}. In \bibinfo{booktitle}{\emph{ESEC/FSE'19: The ACM Joint
  European Software Engineering Conference and Symposium on the Foundations of
  Software Engineering (ESEC/FSE)}} \emph{(\bibinfo{series}{ESEC/FSE 2019})}.
\newblock


\bibitem[\protect\citeauthoryear{Islam, Pan, Nguyen, and Rajan}{Islam
  et~al\mbox{.}}{2020}]%
        {islam20repairing}
\bibfield{author}{\bibinfo{person}{Md~Johirul Islam}, \bibinfo{person}{Rangeet
  Pan}, \bibinfo{person}{Giang Nguyen}, {and} \bibinfo{person}{Hridesh Rajan}.}
  \bibinfo{year}{2020}\natexlab{}.
\newblock \showarticletitle{Repairing Deep Neural Networks: Fix Patterns and
  Challenges}. In \bibinfo{booktitle}{\emph{ICSE'20: The 42nd International
  Conference on Software Engineering}} (Seoul, South Korea).
\newblock


\bibitem[\protect\citeauthoryear{{Kaggle}}{{Kaggle}}{2017a}]%
        {home}
\bibfield{author}{\bibinfo{person}{{Kaggle}}.}
  \bibinfo{year}{2017}\natexlab{a}.
\newblock \bibinfo{title}{{Home Credit Dataset}}.
\newblock
\newblock
\newblock
\shownote{\url{https://www.kaggle.com/c/home-credit-default-risk}.}


\bibitem[\protect\citeauthoryear{{Kaggle}}{{Kaggle}}{2017b}]%
        {titan}
\bibfield{author}{\bibinfo{person}{{Kaggle}}.}
  \bibinfo{year}{2017}\natexlab{b}.
\newblock \bibinfo{title}{{Titanic ML Dataset}}.
\newblock
\newblock
\newblock
\shownote{\url{https://www.kaggle.com/c/titanic/data}.}


\bibitem[\protect\citeauthoryear{Kamiran and Calders}{Kamiran and
  Calders}{2012}]%
        {kamiran2012data}
\bibfield{author}{\bibinfo{person}{Faisal Kamiran} {and} \bibinfo{person}{Toon
  Calders}.} \bibinfo{year}{2012}\natexlab{}.
\newblock \showarticletitle{Data preprocessing techniques for classification
  without discrimination}.
\newblock \bibinfo{journal}{\emph{Knowledge and Information Systems}}
  \bibinfo{volume}{33}, \bibinfo{number}{1} (\bibinfo{year}{2012}),
  \bibinfo{pages}{1--33}.
\newblock


\bibitem[\protect\citeauthoryear{Kamiran, Karim, and Zhang}{Kamiran
  et~al\mbox{.}}{2012}]%
        {kamiran2012decision}
\bibfield{author}{\bibinfo{person}{Faisal Kamiran}, \bibinfo{person}{Asim
  Karim}, {and} \bibinfo{person}{Xiangliang Zhang}.}
  \bibinfo{year}{2012}\natexlab{}.
\newblock \showarticletitle{Decision theory for discrimination-aware
  classification}. In \bibinfo{booktitle}{\emph{2012 IEEE 12th International
  Conference on Data Mining}}. IEEE, \bibinfo{pages}{924--929}.
\newblock


\bibitem[\protect\citeauthoryear{Kamishima, Akaho, Asoh, and Sakuma}{Kamishima
  et~al\mbox{.}}{2012}]%
        {kamishima2012fairness}
\bibfield{author}{\bibinfo{person}{Toshihiro Kamishima},
  \bibinfo{person}{Shotaro Akaho}, \bibinfo{person}{Hideki Asoh}, {and}
  \bibinfo{person}{Jun Sakuma}.} \bibinfo{year}{2012}\natexlab{}.
\newblock \showarticletitle{Fairness-aware classifier with prejudice remover
  regularizer}. In \bibinfo{booktitle}{\emph{Joint European Conference on
  Machine Learning and Knowledge Discovery in Databases}}. Springer,
  \bibinfo{pages}{35--50}.
\newblock


\bibitem[\protect\citeauthoryear{Kirkpatrick}{Kirkpatrick}{2017}]%
        {kirkpatrick2017s}
\bibfield{author}{\bibinfo{person}{Keith Kirkpatrick}.}
  \bibinfo{year}{2017}\natexlab{}.
\newblock \showarticletitle{It's not the algorithm, it's the data}.
\newblock \bibinfo{journal}{\emph{Commun. ACM}} \bibinfo{volume}{60},
  \bibinfo{number}{2} (\bibinfo{year}{2017}), \bibinfo{pages}{21--23}.
\newblock


\bibitem[\protect\citeauthoryear{Kohavi}{Kohavi}{1996}]%
        {kohavi1996scaling}
\bibfield{author}{\bibinfo{person}{Ron Kohavi}.}
  \bibinfo{year}{1996}\natexlab{}.
\newblock \showarticletitle{Scaling up the accuracy of naive-bayes classifiers:
  A decision-tree hybrid.}. In \bibinfo{booktitle}{\emph{KDD}},
  Vol.~\bibinfo{volume}{96}. \bibinfo{pages}{202--207}.
\newblock
\urldef\tempurl%
\url{https://archive.ics.uci.edu/ml/datasets/adult}
\showURL{%
\tempurl}


\bibitem[\protect\citeauthoryear{Kunft, Katsifodimos, Schelter, Bre{\ss}, Rabl,
  and Markl}{Kunft et~al\mbox{.}}{2019}]%
        {kunft2019intermediate}
\bibfield{author}{\bibinfo{person}{Andreas Kunft}, \bibinfo{person}{Asterios
  Katsifodimos}, \bibinfo{person}{Sebastian Schelter},
  \bibinfo{person}{Sebastian Bre{\ss}}, \bibinfo{person}{Tilmann Rabl}, {and}
  \bibinfo{person}{Volker Markl}.} \bibinfo{year}{2019}\natexlab{}.
\newblock \showarticletitle{An intermediate representation for optimizing
  machine learning pipelines}.
\newblock \bibinfo{journal}{\emph{Proceedings of the VLDB Endowment}}
  \bibinfo{volume}{12}, \bibinfo{number}{11} (\bibinfo{year}{2019}),
  \bibinfo{pages}{1553--1567}.
\newblock


\bibitem[\protect\citeauthoryear{Kusner, Loftus, Russell, and Silva}{Kusner
  et~al\mbox{.}}{2017}]%
        {kusner2017counterfactual}
\bibfield{author}{\bibinfo{person}{Matt Kusner}, \bibinfo{person}{Joshua
  Loftus}, \bibinfo{person}{Chris Russell}, {and} \bibinfo{person}{Ricardo
  Silva}.} \bibinfo{year}{2017}\natexlab{}.
\newblock \showarticletitle{Counterfactual fairness}. In
  \bibinfo{booktitle}{\emph{Proceedings of the 31st International Conference on
  Neural Information Processing Systems}}. \bibinfo{pages}{4069--4079}.
\newblock


\bibitem[\protect\citeauthoryear{Lema{\^\i}tre, Nogueira, and
  Aridas}{Lema{\^\i}tre et~al\mbox{.}}{2017}]%
        {lemaitre2017imbalanced}
\bibfield{author}{\bibinfo{person}{Guillaume Lema{\^\i}tre},
  \bibinfo{person}{Fernando Nogueira}, {and} \bibinfo{person}{Christos~K
  Aridas}.} \bibinfo{year}{2017}\natexlab{}.
\newblock \showarticletitle{Imbalanced-learn: A python toolbox to tackle the
  curse of imbalanced datasets in machine learning}.
\newblock \bibinfo{journal}{\emph{The Journal of Machine Learning Research}}
  \bibinfo{volume}{18}, \bibinfo{number}{1} (\bibinfo{year}{2017}),
  \bibinfo{pages}{559--563}.
\newblock


\bibitem[\protect\citeauthoryear{Moro, Cortez, and Rita}{Moro
  et~al\mbox{.}}{2014}]%
        {moro2014data}
\bibfield{author}{\bibinfo{person}{S{\'e}rgio Moro}, \bibinfo{person}{Paulo
  Cortez}, {and} \bibinfo{person}{Paulo Rita}.}
  \bibinfo{year}{2014}\natexlab{}.
\newblock \showarticletitle{A data-driven approach to predict the success of
  bank telemarketing}.
\newblock \bibinfo{journal}{\emph{Decision Support Systems}}
  \bibinfo{volume}{62} (\bibinfo{year}{2014}), \bibinfo{pages}{22--31}.
\newblock
\urldef\tempurl%
\url{https://archive.ics.uci.edu/ml/datasets/Bank+Marketing}
\showURL{%
\tempurl}


\bibitem[\protect\citeauthoryear{Olson}{Olson}{2011}]%
        {olson2011algorithm}
\bibfield{author}{\bibinfo{person}{P Olson}.} \bibinfo{year}{2011}\natexlab{}.
\newblock \showarticletitle{The algorithm that beats your bank manager}.
\newblock \bibinfo{journal}{\emph{CNN Money}} (\bibinfo{year}{2011}).
\newblock
\urldef\tempurl%
\url{https://www.forbes.com/sites/parmyolson/2011/03/15/the-algorithm-that-beats-your-bank-manager/}
\showURL{%
\tempurl}


\bibitem[\protect\citeauthoryear{Olson and Moore}{Olson and Moore}{2016}]%
        {olson2016tpot}
\bibfield{author}{\bibinfo{person}{Randal~S Olson} {and}
  \bibinfo{person}{Jason~H Moore}.} \bibinfo{year}{2016}\natexlab{}.
\newblock \showarticletitle{TPOT: A tree-based pipeline optimization tool for
  automating machine learning}. In \bibinfo{booktitle}{\emph{Workshop on
  automatic machine learning}}. PMLR, \bibinfo{pages}{66--74}.
\newblock


\bibitem[\protect\citeauthoryear{Pan and Rajan}{Pan and Rajan}{2020}]%
        {pan20decomposing}
\bibfield{author}{\bibinfo{person}{Rangeet Pan} {and} \bibinfo{person}{Hridesh
  Rajan}.} \bibinfo{year}{2020}\natexlab{}.
\newblock \showarticletitle{On Decomposing a Deep Neural Network into Modules}.
  In \bibinfo{booktitle}{\emph{ESEC/FSE'2020: The 28th ACM Joint European
  Software Engineering Conference and Symposium on the Foundations of Software
  Engineering}} (Sacramento, California, United States).
\newblock


\bibitem[\protect\citeauthoryear{Pearl}{Pearl}{2000}]%
        {pearl2000causality}
\bibfield{author}{\bibinfo{person}{Judea Pearl}.}
  \bibinfo{year}{2000}\natexlab{}.
\newblock \showarticletitle{Causality: Models, reasoning and inference
  cambridge university press}.
\newblock \bibinfo{journal}{\emph{Cambridge, MA, USA,}}  \bibinfo{volume}{9}
  (\bibinfo{year}{2000}), \bibinfo{pages}{10--11}.
\newblock


\bibitem[\protect\citeauthoryear{Pearl}{Pearl}{2009}]%
        {pearl2009causal}
\bibfield{author}{\bibinfo{person}{Judea Pearl}.}
  \bibinfo{year}{2009}\natexlab{}.
\newblock \showarticletitle{Causal inference in statistics: An overview}.
\newblock \bibinfo{journal}{\emph{Statistics surveys}}  \bibinfo{volume}{3}
  (\bibinfo{year}{2009}), \bibinfo{pages}{96--146}.
\newblock


\bibitem[\protect\citeauthoryear{Pleiss, Raghavan, Wu, Kleinberg, and
  Weinberger}{Pleiss et~al\mbox{.}}{2017}]%
        {pleiss2017fairness}
\bibfield{author}{\bibinfo{person}{Geoff Pleiss}, \bibinfo{person}{Manish
  Raghavan}, \bibinfo{person}{Felix Wu}, \bibinfo{person}{Jon Kleinberg}, {and}
  \bibinfo{person}{Kilian~Q Weinberger}.} \bibinfo{year}{2017}\natexlab{}.
\newblock \showarticletitle{On fairness and calibration}. In
  \bibinfo{booktitle}{\emph{Advances in Neural Information Processing
  Systems}}. \bibinfo{pages}{5680--5689}.
\newblock


\bibitem[\protect\citeauthoryear{Ribeiro, Singh, and Guestrin}{Ribeiro
  et~al\mbox{.}}{2018}]%
        {ribeiro2018anchors}
\bibfield{author}{\bibinfo{person}{Marco~Tulio Ribeiro},
  \bibinfo{person}{Sameer Singh}, {and} \bibinfo{person}{Carlos Guestrin}.}
  \bibinfo{year}{2018}\natexlab{}.
\newblock \showarticletitle{Anchors: High-precision model-agnostic
  explanations}. In \bibinfo{booktitle}{\emph{Proceedings of the AAAI
  Conference on Artificial Intelligence}}, Vol.~\bibinfo{volume}{32}.
\newblock


\bibitem[\protect\citeauthoryear{Russell, Kusner, Loftus, and Silva}{Russell
  et~al\mbox{.}}{2017}]%
        {russell2017worlds}
\bibfield{author}{\bibinfo{person}{Christopher Russell},
  \bibinfo{person}{Matt~J Kusner}, \bibinfo{person}{Joshua~R Loftus}, {and}
  \bibinfo{person}{Ricardo Silva}.} \bibinfo{year}{2017}\natexlab{}.
\newblock \showarticletitle{When worlds collide: integrating different
  counterfactual assumptions in fairness}.
\newblock \bibinfo{journal}{\emph{Advances in Neural Information Processing
  Systems 30. Pre-proceedings}}  \bibinfo{volume}{30} (\bibinfo{year}{2017}).
\newblock


\bibitem[\protect\citeauthoryear{Saleiro, Kuester, Hinkson, London, Stevens,
  Anisfeld, Rodolfa, and Ghani}{Saleiro et~al\mbox{.}}{2018}]%
        {saleiro2018aequitas}
\bibfield{author}{\bibinfo{person}{Pedro Saleiro}, \bibinfo{person}{Benedict
  Kuester}, \bibinfo{person}{Loren Hinkson}, \bibinfo{person}{Jesse London},
  \bibinfo{person}{Abby Stevens}, \bibinfo{person}{Ari Anisfeld},
  \bibinfo{person}{Kit~T Rodolfa}, {and} \bibinfo{person}{Rayid Ghani}.}
  \bibinfo{year}{2018}\natexlab{}.
\newblock \showarticletitle{Aequitas: A bias and fairness audit toolkit}.
\newblock \bibinfo{journal}{\emph{arXiv preprint arXiv:1811.05577}}
  (\bibinfo{year}{2018}).
\newblock


\bibitem[\protect\citeauthoryear{Salimi, Rodriguez, Howe, and Suciu}{Salimi
  et~al\mbox{.}}{2019}]%
        {salimi2019interventional}
\bibfield{author}{\bibinfo{person}{Babak Salimi}, \bibinfo{person}{Luke
  Rodriguez}, \bibinfo{person}{Bill Howe}, {and} \bibinfo{person}{Dan Suciu}.}
  \bibinfo{year}{2019}\natexlab{}.
\newblock \showarticletitle{Interventional fairness: Causal database repair for
  algorithmic fairness}. In \bibinfo{booktitle}{\emph{Proceedings of the 2019
  International Conference on Management of Data}}. \bibinfo{pages}{793--810}.
\newblock


\bibitem[\protect\citeauthoryear{{Scikit Learn}}{{Scikit Learn}}{2019a}]%
        {featselect}
\bibfield{author}{\bibinfo{person}{{Scikit Learn}}.}
  \bibinfo{year}{2019}\natexlab{a}.
\newblock \bibinfo{title}{{Feature Selection Methods}}.
\newblock
\newblock
\newblock
\shownote{\url{https://scikit-learn.org/stable/modules/feature_selection.html}.}


\bibitem[\protect\citeauthoryear{{Scikit Learn}}{{Scikit Learn}}{2019b}]%
        {prepsklearn}
\bibfield{author}{\bibinfo{person}{{Scikit Learn}}.}
  \bibinfo{year}{2019}\natexlab{b}.
\newblock \bibinfo{title}{{Preprocessing API Documentation}}.
\newblock
\newblock
\newblock
\shownote{\url{https://scikit-learn.org/stable/modules/classes.html\#module-sklearn.preprocessing}.}


\bibitem[\protect\citeauthoryear{{Scikit Learn}}{{Scikit Learn}}{2019c}]%
        {imputer}
\bibfield{author}{\bibinfo{person}{{Scikit Learn}}.}
  \bibinfo{year}{2019}\natexlab{c}.
\newblock \bibinfo{title}{{Scikit Learn SimpleImputer}}.
\newblock
\newblock
\newblock
\shownote{\url{https://scikit-learn.org/0.18/modules/generated/sklearn.preprocessing.Imputer.html}.}


\bibitem[\protect\citeauthoryear{{Scikit-Learn Pipeline}}{{Scikit-Learn
  Pipeline}}{2020}]%
        {sklearn-pipeline}
\bibfield{author}{\bibinfo{person}{{Scikit-Learn Pipeline}}.}
  \bibinfo{year}{2020}\natexlab{}.
\newblock \bibinfo{title}{{Scikit-Learn API Documentation}}.
\newblock
\newblock
\newblock
\shownote{\url{https://scikit-learn.org/stable/modules/generated/sklearn.pipeline.Pipeline.html}.}


\bibitem[\protect\citeauthoryear{Sokol, Santos-Rodriguez, and Flach}{Sokol
  et~al\mbox{.}}{2019}]%
        {sokol2019fat}
\bibfield{author}{\bibinfo{person}{Kacper Sokol}, \bibinfo{person}{Raul
  Santos-Rodriguez}, {and} \bibinfo{person}{Peter Flach}.}
  \bibinfo{year}{2019}\natexlab{}.
\newblock \showarticletitle{FAT Forensics: A Python Toolbox for Algorithmic
  Fairness, Accountability and Transparency}.
\newblock \bibinfo{journal}{\emph{arXiv preprint arXiv:1909.05167}}
  (\bibinfo{year}{2019}).
\newblock


\bibitem[\protect\citeauthoryear{Speicher, Heidari, Grgic-Hlaca, Gummadi,
  Singla, Weller, and Zafar}{Speicher et~al\mbox{.}}{2018}]%
        {speicher2018unified}
\bibfield{author}{\bibinfo{person}{Till Speicher}, \bibinfo{person}{Hoda
  Heidari}, \bibinfo{person}{Nina Grgic-Hlaca}, \bibinfo{person}{Krishna~P
  Gummadi}, \bibinfo{person}{Adish Singla}, \bibinfo{person}{Adrian Weller},
  {and} \bibinfo{person}{Muhammad~Bilal Zafar}.}
  \bibinfo{year}{2018}\natexlab{}.
\newblock \showarticletitle{A unified approach to quantifying algorithmic
  unfairness: Measuring individual \&group unfairness via inequality indices}.
  In \bibinfo{booktitle}{\emph{Proceedings of the 24th ACM SIGKDD International
  Conference on Knowledge Discovery \& Data Mining}}.
  \bibinfo{pages}{2239--2248}.
\newblock


\bibitem[\protect\citeauthoryear{Tramer, Atlidakis, Geambasu, Hsu, Hubaux,
  Humbert, Juels, and Lin}{Tramer et~al\mbox{.}}{2017}]%
        {tramer2017fairtest}
\bibfield{author}{\bibinfo{person}{Florian Tramer}, \bibinfo{person}{Vaggelis
  Atlidakis}, \bibinfo{person}{Roxana Geambasu}, \bibinfo{person}{Daniel Hsu},
  \bibinfo{person}{Jean-Pierre Hubaux}, \bibinfo{person}{Mathias Humbert},
  \bibinfo{person}{Ari Juels}, {and} \bibinfo{person}{Huang Lin}.}
  \bibinfo{year}{2017}\natexlab{}.
\newblock \showarticletitle{FairTest: Discovering unwarranted associations in
  data-driven applications}. In \bibinfo{booktitle}{\emph{2017 IEEE European
  Symposium on Security and Privacy (EuroS\&P)}}. IEEE,
  \bibinfo{pages}{401--416}.
\newblock


\bibitem[\protect\citeauthoryear{Udeshi, Arora, and Chattopadhyay}{Udeshi
  et~al\mbox{.}}{2018}]%
        {udeshi2018automated}
\bibfield{author}{\bibinfo{person}{Sakshi Udeshi}, \bibinfo{person}{Pryanshu
  Arora}, {and} \bibinfo{person}{Sudipta Chattopadhyay}.}
  \bibinfo{year}{2018}\natexlab{}.
\newblock \showarticletitle{Automated directed fairness testing}. In
  \bibinfo{booktitle}{\emph{Proceedings of the 33rd ACM/IEEE International
  Conference on Automated Software Engineering}}. \bibinfo{pages}{98--108}.
\newblock


\bibitem[\protect\citeauthoryear{{US Equal Employment Opportunity
  Commission}}{{US Equal Employment Opportunity Commission}}{1979}]%
        {usemploy}
\bibfield{author}{\bibinfo{person}{{US Equal Employment Opportunity
  Commission}}.} \bibinfo{year}{1979}\natexlab{}.
\newblock \bibinfo{title}{{Guidelines on Employee Selection Procedures}}.
\newblock
\newblock
\newblock
\shownote{\url{https://www.eeoc.gov/laws/guidance/questions-and-answers-clarify-and-provide-common-interpretation-uniform-guidelines}.}


\bibitem[\protect\citeauthoryear{Uysal and Gunal}{Uysal and Gunal}{2014}]%
        {uysal2014impact}
\bibfield{author}{\bibinfo{person}{Alper~Kursat Uysal} {and}
  \bibinfo{person}{Serkan Gunal}.} \bibinfo{year}{2014}\natexlab{}.
\newblock \showarticletitle{The impact of preprocessing on text
  classification}.
\newblock \bibinfo{journal}{\emph{Information Processing \& Management}}
  \bibinfo{volume}{50}, \bibinfo{number}{1} (\bibinfo{year}{2014}),
  \bibinfo{pages}{104--112}.
\newblock


\bibitem[\protect\citeauthoryear{Wardat, Le, and Rajan}{Wardat
  et~al\mbox{.}}{2021}]%
        {wardat21deeplocalize}
\bibfield{author}{\bibinfo{person}{Mohammad Wardat}, \bibinfo{person}{Wei Le},
  {and} \bibinfo{person}{Hridesh Rajan}.} \bibinfo{year}{2021}\natexlab{}.
\newblock \showarticletitle{DeepLocalize: Fault Localization for Deep Neural
  Networks}. In \bibinfo{booktitle}{\emph{ICSE'21: The 43nd International
  Conference on Software Engineering}} (Virtual Conference).
\newblock


\bibitem[\protect\citeauthoryear{Yang, Huang, Stoyanovich, and Schelter}{Yang
  et~al\mbox{.}}{2020}]%
        {yang2020fairness}
\bibfield{author}{\bibinfo{person}{Ke Yang}, \bibinfo{person}{Biao Huang},
  \bibinfo{person}{Julia Stoyanovich}, {and} \bibinfo{person}{Sebastian
  Schelter}.} \bibinfo{year}{2020}\natexlab{}.
\newblock \showarticletitle{Fairness-Aware Instrumentation of Preprocessing
  Pipelines for Machine Learning}. In \bibinfo{booktitle}{\emph{Workshop on
  Human-In-the-Loop Data Analytics (HILDA'20)}}.
\newblock


\bibitem[\protect\citeauthoryear{Zafar, Valera, Rodriguez, and Gummadi}{Zafar
  et~al\mbox{.}}{2015}]%
        {zafar2015fairness}
\bibfield{author}{\bibinfo{person}{Muhammad~Bilal Zafar},
  \bibinfo{person}{Isabel Valera}, \bibinfo{person}{Manuel~Gomez Rodriguez},
  {and} \bibinfo{person}{Krishna~P Gummadi}.} \bibinfo{year}{2015}\natexlab{}.
\newblock \showarticletitle{Fairness constraints: Mechanisms for fair
  classification}.
\newblock \bibinfo{journal}{\emph{arXiv preprint arXiv:1507.05259}}
  (\bibinfo{year}{2015}).
\newblock


\bibitem[\protect\citeauthoryear{Zelaya}{Zelaya}{2019}]%
        {zelaya2019towards}
\bibfield{author}{\bibinfo{person}{Carlos Vladimiro~Gonz{\'a}lez Zelaya}.}
  \bibinfo{year}{2019}\natexlab{}.
\newblock \showarticletitle{Towards Explaining the Effects of Data
  Preprocessing on Machine Learning}. In \bibinfo{booktitle}{\emph{2019 IEEE
  35th International Conference on Data Engineering (ICDE)}}. IEEE,
  \bibinfo{pages}{2086--2090}.
\newblock


\bibitem[\protect\citeauthoryear{Zemel, Wu, Swersky, Pitassi, and Dwork}{Zemel
  et~al\mbox{.}}{2013}]%
        {zemel2013learning}
\bibfield{author}{\bibinfo{person}{Rich Zemel}, \bibinfo{person}{Yu Wu},
  \bibinfo{person}{Kevin Swersky}, \bibinfo{person}{Toni Pitassi}, {and}
  \bibinfo{person}{Cynthia Dwork}.} \bibinfo{year}{2013}\natexlab{}.
\newblock \showarticletitle{Learning fair representations}. In
  \bibinfo{booktitle}{\emph{International Conference on Machine Learning}}.
  \bibinfo{pages}{325--333}.
\newblock


\bibitem[\protect\citeauthoryear{Zhang, Lemoine, and Mitchell}{Zhang
  et~al\mbox{.}}{2018}]%
        {zhang2018mitigating}
\bibfield{author}{\bibinfo{person}{Brian~Hu Zhang}, \bibinfo{person}{Blake
  Lemoine}, {and} \bibinfo{person}{Margaret Mitchell}.}
  \bibinfo{year}{2018}\natexlab{}.
\newblock \showarticletitle{Mitigating unwanted biases with adversarial
  learning}. In \bibinfo{booktitle}{\emph{Proceedings of the 2018 AAAI/ACM
  Conference on AI, Ethics, and Society}}. \bibinfo{pages}{335--340}.
\newblock


\bibitem[\protect\citeauthoryear{Zhang and Bareinboim}{Zhang and
  Bareinboim}{2018}]%
        {zhang2018fairness}
\bibfield{author}{\bibinfo{person}{Junzhe Zhang} {and} \bibinfo{person}{Elias
  Bareinboim}.} \bibinfo{year}{2018}\natexlab{}.
\newblock \showarticletitle{Fairness in decision-making—the causal
  explanation formula}. In \bibinfo{booktitle}{\emph{Proceedings of the AAAI
  Conference on Artificial Intelligence}}, Vol.~\bibinfo{volume}{32}.
\newblock


\bibitem[\protect\citeauthoryear{Zhang and Harman}{Zhang and Harman}{2021}]%
        {zhang2021ignorance}
\bibfield{author}{\bibinfo{person}{Jie~M Zhang} {and} \bibinfo{person}{Mark
  Harman}.} \bibinfo{year}{2021}\natexlab{}.
\newblock \showarticletitle{“Ignorance and Prejudice” in Software
  Fairness}. In \bibinfo{booktitle}{\emph{2021 IEEE/ACM 43rd International
  Conference on Software Engineering (ICSE)}}. IEEE,
  \bibinfo{pages}{1436--1447}.
\newblock


\end{thebibliography}

\end{document}